%% file: main.tex
\documentclass[12pt]{graphicsclass}
\usepackage[utf8]{inputenc}
\usepackage{palatino}
\usepackage{amsmath}
\usepackage{amssymb} 
\usepackage{graphicx}
\graphicspath{ {images/} }
\usepackage[labelfont=bf]{caption}
\usepackage{subcaption}
\usepackage{setspace}
\usepackage[margin=1in]{geometry}
\usepackage[hidelinks]{hyperref}
\usepackage{cite}
\usepackage{overpic}
\usepackage[all]{nowidow}
\usepackage{algorithm, algpseudocode}
\usepackage{tabularx}
\usepackage{titlesec}
\usepackage{float}
\usepackage{enumerate}
\usepackage{booktabs}
\usepackage{xspace}

\usepackage[inline]{enumitem}

\usepackage{adjustbox}
\usepackage{longtable} 
\usepackage[table]{xcolor}
\definecolor{p}{RGB}{136,10,255}

\usepackage{graphicx}
\definecolor{gold}{rgb}{1.00,0.84,0.00}
\definecolor{silver}{rgb}{0.75,0.75,0.75}
\definecolor{bronze}{rgb}{0.86,0.67,0.47}
\usepackage{multirow}
\usepackage{diagbox}

\title{
{Representation Integrity in Temporal Graph Learning Methods}\\~\\~\\
{\large Elahe Kooshafar, School of Computer Science \\ 
	McGill University, Montreal \\ 
	July, 2025 \\~\\~\\
	A thesis submitted to McGill University in partial fulfillment of the requirements of the degree of \\~\\ Master of Computer Science }\\~\\
}

\author{\textcopyright Elahe Kooshafar, July 2025}
\date{}

\begin{document}
\maketitle

\chapter*{Abstract}
\label{sec:engAbstract}
\addcontentsline{toc}{section}{\nameref{sec:engAbstract}}

Real-world systems ranging from airline routes to cryptocurrency transfers
are naturally modelled as \emph{dynamic graphs} whose topology changes over
time.  Conventional benchmarks judge dynamic--graph learners by a handful of
task-specific scores, yet seldom ask whether the embeddings themselves
remain a truthful, interpretable reflection of the evolving network.
We formalize this requirement as \textbf{representation integrity} and
derive a family of indexes that measure how closely embedding
changes follow graph changes.  Three synthetic scenarios---\emph{Gradual
Merge}, \emph{Abrupt Move}, and \emph{Periodic Re--wiring}---are used to screen forty--two candidate indexes. Based on which we recommend one index that passes all of our theoretical and empirical tests. In particular, this validated metric consistently ranks
the provably stable UASE and IPP models highest. We then use this index to do a comparative study on representation integrity of common dynamic graph learning models. This study exposes the scenario--specific strengths of neural methods, and shows a strong positive rank correlation
with one-step link-prediction AUC. 
The proposed integrity framework, therefore, offers a task-agnostic and interpretable evaluation tool for dynamic-graph representation quality, providing more explicit guidance for model selection and future architecture design.

\chapter*{Abrégé}
\label{sec:frAbstract}
\addcontentsline{toc}{section}{\nameref{sec:frAbstract}}

Les systèmes réels, allant des itinéraires aériens aux transferts de cryptomonnaies, sont naturellement modélisés comme des graphes dynamiques dont la topologie évolue au fil du temps. Les benchmarks conventionnels évaluent les apprenants de graphes dynamiques selon quelques scores spécifiques à chaque tâche, mais se demandent rarement si les représentations elles-mêmes restent un reflet fidèle et interprétable de l'évolution du réseau.
Nous formalisons cette exigence par l'intégrité de la représentation et dérivons une famille d'indices qui mesurent la proximité entre les variations de représentation et les variations du graphe. Trois scénarios synthétiques : Fusion graduelle, Déplacement brusque et Recâblage périodique, sont utilisés pour sélectionner quarante-deux indices candidats. Sur cette base, nous recommandons un indice qui réussit tous nos tests théoriques et empiriques. En particulier, cette mesure validée classe systématiquement les modèles UASE et IPP dont la stabilité est prouvée en tête. Nous utilisons ensuite cet indice pour réaliser une étude comparative de l'intégrité de la représentation des modèles courants d'apprentissage de graphes dynamiques. Cette étude expose les atouts des méthodes neuronales, spécifiques à chaque scénario, et montre une forte corrélation positive des rangs avec l'AUC de prédiction de lien en une étape. 
Le cadre d'intégrité proposé offre donc un outil d'évaluation de la qualité de la représentation des graphes dynamiques, indépendant des tâches et interprétable, fournissant des indications plus explicites pour la sélection des modèles et la conception des architectures futures.

\chapter*{Acknowledgements}
\label{sec:ded}
\addcontentsline{toc}{section}{\nameref{sec:ded}}

I would like to express my deepest gratitude to my advisor, Professor Reihaneh Rabbany, for her time and efforts in guiding this work, reviewing this thesis, and supporting my research throughout my graduate studies.

I am also deeply thankful to my friends and family for their encouragement and support through every challenge. This research would not have been possible without you.

\tableofcontents
\listoftables
\addcontentsline{toc}{section}{\listtablename}

\clearpage 
\pagenumbering{arabic} 

\input{Chapters/introduction.tex}

\input{Chapters/relatedwork.tex}

\input{Chapters/methodology.tex}
\input{Chapters/experiments.tex}

\input{Chapters/conclusion.tex}

\appendix

\input{Chapters/appendix.tex}

\bibliography{references}
\bibliographystyle{acm}

\end{document}

%% file: Chapters/introduction.tex
\chapter{Introduction}


Many real-world networks-such as transportation systems, trade networks, and cryptocurrency networks-evolve over time. When the graph structure changes across time, the network is modeled as a dynamic graph. As deep learning and network science continue to advance, dynamic graph learning has become essential for analyzing such complex, time-varying systems \cite{JMLR:v21:19-447}. Representation integrity refers to how truthfully the learned representations reflect both the structural and semantic patterns of the input data, independent of any specific downstream task. This thesis focuses on assessing representation integrity in dynamic graph learning methods \cite{9439502}, emphasizing its role in capturing reliable and meaningful network dynamics.

Representation integrity encompasses both trustworthiness and interpretability, serving as a foundation for understanding the quality and reliability of learned representations \cite{davis2023simplepowerfulframeworkstable, 10.5555/3540261.3541038}. We identify two key aspects of representation integrity in dynamic graph models: Static Integrity, which evaluates how well node embeddings at a given time reflect the graph’s structure and semantics, and Temporal Integrity, which assesses how faithfully changes in node embeddings over time reflect genuine structural and semantic changes in the graph, such as shifts in behavioral patterns.

In this work, we propose indexes to quantitatively measure Temporal Integrity in dynamic graph learning models. This framework enables interpretable and diagnostic assessment at the representation level rather than relying solely on specific downstream tasks. It provides a complementary evaluation tool to standard metrics for temporal graph learning models, addressing some evaluation challenges identified in recent studies \cite{Poursafaei2022TowardsBE, bechlerspeicher2025positiongraphlearninglose}.

Improving the truthfulness of dynamic graph representations has practical implications across a range of domains. In anomaly detection, models with high representation integrity can more reliably identify unusual changes in node behavior or relationships, which is critical in cybersecurity and fraud detection. In trend analysis-such as within social networks or financial markets-temporally consistent embeddings help track evolving patterns with greater accuracy. Additionally, in predictive tasks like link prediction or node classification, high-integrity embeddings yield more interpretable and consistent outputs over time. Enhanced temporal coherence in such embeddings can lead to better detection of behavioral shifts and more accurate forecasting. By focusing on temporal representation integrity we enable more robust and meaningful analysis of dynamic networks.

By focusing on both dynamic and static aspects of representation integrity, researchers can develop more robust and reliable dynamic graph representation learning models. Future research should also expand this evaluation framework across a wider range of models and datasets to further explore the broader implications of representation integrity. This research aims to pave the way for more robust and reliable analyses of complex, evolving systems, ultimately enhancing our ability to understand and predict the behavior of dynamic networks across various domains.


\section{Outline of the Thesis}\label{sec:outline}
Chapter 2 surveys the theoretical background-graph theory, static and dynamic embedding methods, and earlier notions of stability-to show why integrity is a natural next step. Chapter 3 formalizes temporal integrity, proposes a modular metric design space that combines graph-change and representation-change measures with an alignment kernel, states the properties a sound index must satisfy, and details three synthetic stress-test scenarios (Gradual Merge, Abrupt Move, Periodic Re-wiring). Chapter 4 validates forty-two candidate indexes against these properties and scenarios, selects the best metric, and deploys it to compare eight representative dynamic-embedding models on both synthetic and real data, also examining how integrity correlates with link-prediction performance. Finally, Chapter 5 interprets the empirical results, highlights scenario-specific strengths and weaknesses, and outlines avenues for future work, while an appendix presents complete numerical tables for transparency.

\section{Statement of Contributions}\label{sec:contribs}
This work makes four main contributions to temporal graph representation learning:

\begin{itemize}
\item We introduces \emph{representation integrity} as a precise, task-agnostic notion of truthfulness for dynamic embeddings, separating static and temporal quality.
\item We propose a modular evaluation framework that factors an integrity index into a graph-change measure, a representation-change measure, and an alignment kernel, supported by formal properties.
\item We screen forty-two candidate indexes and identify the \emph{Euclidean graph change, Euclidean representation change, Pearson alignment} metric as one satisfying all properties across all stress tests.
\item We provide the first comprehensive integrity-based comparison of eight dynamic-embedding models on controlled synthetic scenarios and a real parliamentary-voting dataset, revealing clear scenario-specific performance patterns and a strong positive correlation between integrity and downstream link-prediction accuracy.
\end{itemize}

%% file: Chapters/relatedwork.tex
\chapter{Background and Literature Review}

\section{Graphs}
Graphs are a powerful data structure used to represent complex relationships between entities. A graph consists of a set of nodes (also called vertices) and a set of edges connecting pairs of nodes. In a graph, nodes represent entities, while edges represent the connections between these entities.\cite{10.1093/acprof:oso/9780199206650.001.0001}. This structure is particularly useful in various domains such as social networks \cite{doi:10.1080/23311983.2016.1171458}, biological networks \cite{alma9972283310208496}, transportation systems \cite{Li2024GraphNN}, and citation networks \cite{Li2023AMG}. For instance, in social networks, nodes can represent individuals, and edges can represent friendships or interactions between them.

\subsection{Static Graphs}

A static graph is a type of graph in which the set of nodes and the set of edges connecting these nodes are fixed and do not change over time. Formally, a static graph is represented as $G = (V, E)$ where $V = \{v_1, v_2, . . . , v_{|V|}\}$ is the set of vertices
and $E \subseteq V \times V$ is the set of edges \cite{west01}. 

\subsubsection{Stochastic Block Model}
The Stochastic Block Model (SBM) is a generative model to create synthetic graphs that often exhibit community structure \cite{Holland1983StochasticBF}. In this model, each node belongs to a specific community, and the likelihood of an edge existing between any two nodes is determined by their community memberships. The SBM is characterized by several parameters: the number of nodes \( n \), the number of communities \( C \), a probability vector \( \alpha = (\alpha_1, \alpha_2, \ldots, \alpha_C) \) that defines the distribution of nodes across the communities, and a symmetric matrix \( W \in \mathbb{R}^{C \times C} \) with entries in the range \([0, 1]\), which specifies the probabilities of connections between nodes in different communities.

A graph \( G \) with community labels \( X \) is generated under the SBM model \( \text{SBM}(n, \alpha, W) \) as follows: the community label vector \( X \) is an \( n \)-dimensional vector where each component is independently drawn according to the distribution \( \alpha \). The graph \( G \) is then constructed by connecting any two nodes \( i \) and \( j \) with a probability \( W_{X_i, X_j} \), independently of other pairs of nodes.

\subsection{Dynamic Graphs}
Traditionally, research in graph theory and graph-based machine learning has concentrated on static graphs, where the structure remains unchanged over time \cite{e3da81eedd1340ac81c098b22fe0df41}. However, many real-world systems are inherently dynamic, characterized by evolving nodes and edges that change continuously or discretely over time \cite{e3da81eedd1340ac81c098b22fe0df41}. In the literature, the terms "dynamic" and "temporal" are often used interchangeably to describe evolving graphs or data over time. In this work, we will follow the same convention and use both terms interchangeably.

Dynamic graphs are essential for understanding complex systems in fields such as social networks, communication networks, biological systems, and recommender systems \cite{Holme_2012}. For instance, in social media, communication events such as emails and text messages are
streaming while friendship relations evolve. In recommender systems, new products, new users and new ratings appear every day. 

There are two primary forms of dynamic graphs: discrete-time dynamic graphs (DTDGs) and continuous-time dynamic graphs (CTDGs) \cite{JMLR:v21:19-447}.

\begin{itemize} 
    \item A discrete-time dynamic graph (DTDG) is a series of graph snapshots taken at regular intervals. Formally, we define a DTDG as a set \(\{G_1, G_2, \ldots, G_T\}\), where \(G_t = \{V_t, E_t\}\) represents the graph at snapshot \(t\). Here, \(V_t\) is the set of nodes, and \(E_t\) is the set of edges at time \(t\). The term dynamic graph encompasses both discrete-time dynamic graph (DTDGs) and continuous-time dynamic graphs (CTDGs).

    \item A continuous-time dynamic graph (CTDG) can be represented as a pair \((G, O)\), where \(G\) denotes a static graph that captures the initial state of the dynamic graph at time \(t_0\), and \(O\) is a set of observations or events. Each observation in \(O\) is a tuple of the event's information and timestamp. The event types can include actions such as edge addition, edge deletion, node addition, node deletion, node splitting, node merging, and others.
\end{itemize}

\section{Graph Representation Learning}

Before the advent of deep learning, graph representation primarily relied on traditional methods such as adjacency matrices, feature extraction \cite{2011snda.book..115B, doi:10.1126/science.290.5500.2323}, Laplacian eigenmaps \cite{Belkin2003LaplacianEF}, and spectral clustering \cite{d69168735a23434b90720e92b752c533}. While these approaches laid the foundation for graph analysis, they often struggled with scalability and comprehensive information capture, especially for large and complex graphs \cite{10.1145/3633518}. Specifically, these methods faced challenges in processing graphs with millions of nodes and edges, and often failed to capture higher-order structural information \cite{Hamilton2017RepresentationLO}.

In order to use graphs in different downstream applications, it is important to represent them effectively. Advancements in data availability and computational resources have led to the development of more flexible graph representation methods, particularly graph embedding techniques. These newer approaches, including node2vec \cite{10.1145/2939672.2939754}, DeepWalk \cite{10.1145/2623330.2623732}, and Graph Convolutional Networks (GCNs) \cite{kipf2017semisupervised}, offer improved scalability and the ability to capture complex structural information.

Graph embedding methods project graph elements (nodes, edges, subgraphs) into a lower-dimensional space while preserving important structural and semantic information \cite{Hamilton2017RepresentationLO, 8392745}. Graph embeddings benefit various fields, from computational social science to computational biology \cite{Hamilton2017RepresentationLO}. In the literature, the terms "representation" and "embedding" are often used interchangeably when referring to the mapping of nodes or other graph elements to a vector space. In this work, we will follow the same convention and use both terms interchangeably.

Node representation learning is a prominent part of graph representation learning. It refers to the process of learning to map nodes in a graph to dense vector representations in a low-dimensional space. Node representations, or embeddings, capture the structural and semantic information of the nodes, allowing for efficient computation and analysis of graph data. 

Formally, given a graph $G = (V, E)$, where $V$ is the set of nodes and $E$ is the set of edges, node embedding aims to learn a function $f: V \rightarrow \mathbb{R}^d$ that maps each node $v \in V$ to a $d$-dimensional vector, where $d$ is typically much smaller than $|V|$. Node embeddings have several common applications, including graph visualization, clustering, node classification, and link prediction \cite{Hamilton2017RepresentationLO}.

We will explore relevant works in this area in two subsections: static and dynamic graph representation learning methods. Both categories can be further divided into traditional graph embedding and graph neural network (GNN) based approaches \cite{10.1145/3633518}. Traditional methods capture graph information using techniques such as random walks, matrix factorization, and non-GNN deep learning algorithms \cite{10.1145/3633518}. In contrast, GNN-based methods leverage neural network architectures specifically designed for graph-structured data.

\subsection{Static Graph Representation Learning Methods}

These methods are divided into traditional graph embedding methods and Graph Neural Network(GNN)-based methods. Traditional methods include dimension-reduction-based methods, random walks, factorization methods, and non-GNN deep learning \cite{10.1145/3633518}. 
\begin{itemize}
\item
Factorization-based methods capture the relationships between nodes using a matrix representation and then decompose this matrix to derive node embeddings \cite{Belkin2001LaplacianEA} \cite{doi:10.1126/science.290.5500.2323}. Commonly used matrices include the node adjacency matrix, Laplacian matrix, node transition probability matrix, and Katz similarity matrix, among others. The specific method for matrix factorization depends on the properties of the matrix. For instance, if the matrix is positive semi-definite, like the Laplacian matrix, eigenvalue decomposition can be applied. For more general, unstructured matrices, gradient descent methods can be used to compute the embeddings efficiently in linear time.
\item
Random walk-based methods generate node embeddings by utilizing the statistics of random walks. The key innovation of these approaches is to optimize the embeddings such that nodes frequently appearing together in short random walks over the graph have similar embeddings. In contrast to the deterministic similarity measures used in factorization-based methods, random walk-based approaches rely on a more flexible, stochastic notion of node similarity. Prominent examples of this category include DeepWalk \cite{10.1145/2623330.2623732} and node2vec \cite{10.1145/2939672.2939754}.
\item Non-GNN deep learning methods utilize various techniques outside of GNN architectures to learn embeddings. SDNE \cite{10.1145/2939672.2939753}, for instance, is based on an autoencoder which tries to reconstruct the adjacency
matrix of a graph and captures nodes’ first-order and second-order proximities. Other notable works in this category include \cite{10.5555/3327546.3327592}, \cite{10.1145/3308558.3313668}, and \cite{10.5555/3304222.3304235}.
\end{itemize}

\subsubsection{Graph Neural Networks}
Graph Neural Network (GNN) models represent a category of neural networks specially crafted to process graph data \cite{4700287}. A GNN is a deep learning model which generates a node embedding by aggregating the node’s neighbors embeddings. The GNN’s intuition is that a node’s state is influenced by its interactions with its neighbors in the graph \cite{kipf2017semisupervised,  10.5555/3294771.3294869}. Since their inception, GNNs have seen rapid development in various architectures, applications, and theoretical studies \cite{zhou2020graph}. 

\paragraph{Architecture}
GNNs can generate node representation vectors by stacking several GNN layers. Let $h_i^{l}$ represent the node embeddings for node $i$ at layer $l$. Each GNN layer takes as input the nodes embeddings. The node
representations for node $i$ at each layer $l+1$ are updated using the following formula:
$$h_i^{l+1} = f(h_i^{l}, \sum_{j\in N(i)} g(i, j))$$
where $f$ and $g$ are learnable functions and $N(i)$ are the neighbors of node $i$. $h_i^{0} $ are the initial features of node $i$ \cite{Xu2018HowPA}. At each layer, the embedding of the node $i$ is obtained by aggregating the embeddings of the node’s neighbors. After passing through $L$ GNN layers, the final representation of node $i$ is $h_i^L$, which is the aggregation the node’s neighbors of $L$ hops
away from the node. 

Many GNN models have been developed in recent years. Here, we highlight some key models that have made significant contributions to advancing graph-based learning techniques.

\begin{itemize}
    \item Graph convolution neural network (GCN): GCNs are a type of neural network specifically designed to operate on graph-structured data \cite{kipf2017semisupervised}. They extend the principles of convolutional neural networks (CNNs), which are widely used for grid-like data such as images, to non-Euclidean domains like graphs. GCNs work by performing convolution operations directly on the graph, aggregating information from a node’s neighbors to update its representation. This process, known as neighborhood aggregation or message passing, allows GCNs to capture the local structure and features of the graph effectively.

    For a static graph $G = (V, E)$, where $V$ is the set of nodes and $E$ is the set of edges, the GCN layer can be formulated as \cite{kipf2017semisupervised}:

    \begin{equation}
    H^{(l+1)} = \sigma(\tilde{D}^{-\frac{1}{2}}\tilde{A}\tilde{D}^{-\frac{1}{2}}H^{(l)}W^{(l)})
    \end{equation}

    Where:
    \begin{itemize}
        \item $H^{(l)}$ is the matrix of node features at layer $l$
        \item $\tilde{A} = A + I_N$ is the adjacency matrix with added self-connections ($I_N$ is the identity matrix of size $N\times N$)
        \item $\tilde{D}$ is the degree matrix of $\tilde{A}$
        \item $W^{(l)}$ is the weight matrix for layer $l$
        \item $\sigma(\cdot)$ is a non-linear activation function
    \end{itemize}

    For dynamic graphs, where the graph structure and node features evolve over time, the formulation extends to incorporate temporal dependencies. Given a sequence of graphs $G = \{G^1, G^2, ..., G^T\}$, the dynamic GCN layer at time $t$ can be expressed as \cite{Manessi_2020}:

    \begin{equation}
    H_t^{(l+1)} = \sigma(\tilde{D}_t^{-\frac{1}{2}}A_t\tilde{D}_t^{-\frac{1}{2}}H_t^{(l)}W^{(l)} + H_{t-1}^{(l)}U^{(l)})
    \end{equation}
    
    Where:
    \begin{itemize}
        \item $H_t^{(l)}$ represents the node features at time $t$ and layer $l$
        \item $A_t$ and $\tilde{D}_t$ are the adjacency and degree matrices at time $t$
        \item $W^{(l)}$ is the spatial weight matrix
        \item $U^{(l)}$ is a temporal component (could be a weight matrix or some temporal function).
        \item \( \sigma \) is a non-linear activation function (e.g., ReLU).
    \end{itemize}
    
    This formulation allows the model to capture both spatial dependencies within each graph snapshot and temporal dependencies across different time steps, making it suitable for analyzing evolving graph structures.

    \label{gat}
    \item Graph attention network (GAT): The Graph Attention Network is an advanced neural network designed for graph-structured data, which addresses limitations of earlier methods that relied on graph convolutions or their approximations. Unlike traditional approaches, GAT \cite{veličković2018graph} leverages masked self-attentional layers to address the shortcomings of prior methods. By stacking layers in which nodes are able to attend over their neighborhoods' features, it enables (implicitly) specifying different weights to different nodes in a neighborhood. GATs are well-suited for both inductive and transductive tasks due to their ability to dynamically assign attention weights to neighboring nodes, enabling effective learning from local graph structures and generalization to unseen nodes or edges.

    \item GraphSAGE: GraphSAGE \cite{10.5555/3294771.3294869} is a framework in the field of Graph Neural Networks (GNNs) that addresses the challenges associated with learning on large-scale graphs.

    GraphSAGE operates by first selecting a set number of neighboring nodes for each node to focus on, rather than using the entire graph. It then aggregates information from these selected neighbors, using methods like averaging their features. This aggregated information is combined with the node's own features to produce a new, updated representation for that node. This process enables GraphSAGE to handle large graphs efficiently and to create useful embeddings even for new nodes not seen during training. GraphSAGE allows for various aggregation functions, including mean, LSTM, and pooling operations, to capture different aspects of the node’s local structure.
\end{itemize}

\paragraph{Training Setting}

GNNs can be trained in supervised, semi-supervised or unsupervised frameworks \cite{Zhou2018GraphNN}. In both supervised and semi-supervised frameworks, a variety of prediction tasks, such as node classification and link prediction, can be employed to train the model, targeting nodes, edges, or entire graphs. For example, in node classification within a supervised framework, node embeddings generated by the output of GNN layers are passed through an MLP or softmax layer to produce predictions. A typical loss function for training a GNN in a binary node classification task is the cross-entropy loss, formulated as:
$$L = \sum_{i\in V(train)} y_i \log(\sigma(h_i^{T}\theta)) + (1-y_i) \log(1-\sigma(h_i^{T}\theta))$$
where $h_i$ is the embedding of node $i$ which is the output of the last layer of GNN, $h_i^{T}$ is the transpose of $h_i$, and $y_i$ is the true class label of
the node and $\theta$ are the classification weights.

\paragraph{Downstream Applications}

With the abundance of graph-structured data in science and society, GNNs have found wide applicability, with meaningful impact spanning diverse domains, which encompass recommender systems \cite{10.1145/3539618.3591674}, computer vision \cite{9892658}, natural language processing \cite{10.1609/aaai.v33i01.33017370}, program analysis, software mining, bioinformatics \cite{Li2023AMG}, anomaly detection \cite{2022IEEEA..10k1820K}, and urban intelligence \cite{Li2022}, among others. The fundamental prerequisite for GNN utilization is the transformation or representation of input data into a graph-like structure. In the realm of graph representation learning, GNNs excel in acquiring essential node or graph embeddings that serve as a crucial foundation for subsequent tasks \cite{Zhou2018GraphNN}. The downstream applications where GNNs have proven effective can be categorized as follows:

\begin{itemize}
\item Node classification involves assigning labels to nodes by leveraging the labels of their neighboring nodes. This task is typically approached using semi-supervised learning, where only a subset of the graph is labeled. The algorithm learns from the labeled nodes and generalizes to label the remaining nodes.
\item Graph classification entails categorizing entire graphs into distinct classes, analogous to image classification but applied to graph structures. This method has diverse applications, such as cancer subtype classification in bioinformatics \cite{Li2023AMG} or text classification in natural language processing \cite{10.1609/aaai.v33i01.33017370}.
\item Graph visualization is a field intersecting geometric graph theory and information visualization, focusing on the visual representation of graphs \cite{10.1111:j.1467-8659.2011.01898.x}. Effective graph visualization techniques reveal underlying structures and anomalies within the data, aiding users in comprehending complex graph-based information \cite{zheng2024improving}.
\item Link prediction aims to infer the existence of connections between entities within a graph. This technique is crucial in social networks for deducing potential social interactions or suggesting new connections, such as friends or collaborators \cite{\cite{LI2023126441}}. Additionally, link prediction is applied in recommender systems \cite{10.1145/3539618.3591674} and in identifying potential criminal associations.
\item Graph Clustering is the process of grouping the nodes of the graph into clusters, taking into account the edge structure of the graph in such a way that there are several edges within each cluster and very few between clusters. Graph Clustering intends to partition the nodes in the graph into disjoint groups \cite{Bo2020StructuralDC} \cite{49951}. 
\end{itemize} 

\subsection{Dynamic Graph Representation Learning Methods}

Dynamic graph learning methods are divided into traditional dynamic graph embedding methods and Dynamic Graph Neural Network(DGNN)-based methods. Traditional dynamic graph representation learning approaches can be categorized into four main types: Aggregation based, Random walk based, Non-GNN based deep learning based, and temporal point process based methods. These approaches aim to capture various dynamic behaviors in networks such as topological evolution and temporal link prediction.

\begin{itemize}
\item \textbf{Aggregation based methods:}
Aggregation-based dynamic graph embedding methods generate embeddings by combining information from dynamic graphs. They generally fall into two categories. The first involves aggregating temporal features, where the evolution of the graph is condensed into a single static graph and static graph embedding methods are applied on the single graph to generate the embeddings. Techniques such as summing adjacency matrices or using weighted sums that emphasize more recent snapshots are applied. However, this approach loses detailed temporal information, such as the specific timing of edge creation. Factorization-based models also fit into this category by representing the sequence of graphs as a three-dimensional tensor and applying factorization techniques to generate embeddings.

The second category involves aggregating static embeddings. In this approach, static embedding methods are first applied to each graph snapshot individually. The resulting embeddings are then combined into a single matrix, often using a decay factor to prioritize more recent snapshots. Alternatively, time-series models like ARIMA may be used to predict embeddings for future snapshots based on the sequence of previous graphs \cite{6252471}.
    
\item \textbf{Random walk based methods:}
 These methods extend traditional random walk techniques to dynamic graphs. By simulating random walks over time, they capture the evolving connectivity patterns. The embeddings are updated based on the sequences of nodes visited in these walks, which helps in capturing temporal dependencies \cite{10.1145/3483595}.
    
\item \textbf{Non-GNN based deep learning methods:}
Non-GNN based deep learning methods for dynamic graph embedding, such as DynGEM \cite{goyal2018dyngem}, Dyn-VGAE\cite{Mahdavi2019DynamicJV}, and Dyngraph2vec \cite{Goyal2018dyngraph2vecCN}, utilize models like RNNs and autoencoders to capture temporal graph evolution. DynGEM generates embeddings by initializing each snapshot with the previous time step's embeddings, adapting to graph growth. Dyn-VGAE uses a loss function combining VGAE \cite{Kipf2016VariationalGA} loss and KL divergence to maintain temporal consistency. Dyngraph2vec processes adjacency matrices from past snapshots to generate embeddings, with variants using different network architectures. These methods focus on preserving both structural integrity and temporal patterns as the graph evolves.
    
\item \textbf{Temporal point process based:}
    Temporal point process-based methods are a class of dynamic graph embedding techniques that model the interactions between nodes as stochastic processes occurring over time \cite{10.1145/3483595}. These methods assume that the formation and evolution of graph structures are driven by underlying probabilistic events, which can be captured and analyzed using temporal point processes.

    In this framework, the occurrence of edges between nodes is treated as discrete events happening at specific time points, allowing the model to predict when and how these interactions will occur in the future. By leveraging the rich statistical properties of point processes, these methods can effectively capture the temporal dynamics of node interactions, offering insights into the underlying mechanisms driving graph evolution.
\end{itemize}

\subsubsection{Dynamic Graph Neural Networks}

Dynamic Graph Neural Networks (DGNNs) are designed to manage graphs that change over time. They process sequences of graph snapshots, each representing the graph at a specific moment. For each snapshot, DGNNs create embeddings that reflect both the current state of the graph and its historical changes.

DGNNs employ dynamic message-passing algorithms that not only consider the current graph structure but also integrate information from previous snapshots. This approach allows nodes to exchange information across different time steps.

To capture temporal changes, DGNNs aggregate information from past snapshots. This can involve techniques such as recurrent neural networks (RNNs) or attention mechanisms, which weigh more recent data more heavily. As the graph evolves, DGNNs continuously update node embeddings to reflect both new and historical information, enabling the model to adapt to new patterns and trends.

In addition to processing current data, DGNNs use learned temporal sequences to predict future changes in the graph, such as emerging edges or evolving node attributes.

DGNNs are applied in various fields, including social networks for tracking user interactions, traffic systems for forecasting congestion, financial markets for predicting transaction trends, and biological networks for studying process changes over time. Overall, DGNNs extend traditional graph neural networks by integrating temporal information, allowing them to handle and predict dynamic changes in graphs.

\paragraph{Dynamic Graph Embedding Applications}

Dynamic graph embeddings are used to represent graphs that evolve over time, capturing both structural and temporal changes. These embeddings find applications across various domains, providing insights and predictions based on dynamic processes.

\begin{itemize}
    \item \textbf{Social Networks:} In social networks, dynamic graph embeddings track changes in user interactions and relationships. They help in understanding evolving social dynamics, predicting user behavior \cite{10.1007/s10994-023-06475-x}, and identifying emerging communities or influential nodes.
    
    \item \textbf{Traffic Systems:} For traffic systems, dynamic graph embeddings analyze and forecast traffic flow \cite{8917213} and congestion patterns. They model the changing road network, predict traffic jams, and optimize routing based on historical and real-time data.
    
    \item \textbf{Financial Markets:} In financial markets, these embeddings monitor and predict shifts in transactions and market trends \cite{qian2024mdgnn}. They assist in detecting anomalies, forecasting stock prices, and understanding the flow of financial activities over time.
    
    \item \textbf{Biological Networks:} In biological networks, dynamic graph embeddings study changes in protein interactions, gene expressions, and other biological processes. They help in understanding disease progression, drug interactions, and the temporal evolution of biological systems \cite{quan2024clustering}.
    
    \item \textbf{Recommendation Systems:} Dynamic graph embeddings enhance recommendation systems by adapting to users' changing preferences and interactions \cite{10.1145/3539618.3591674}. They improve recommendations by capturing evolving user-item relationships and trends.
    
    \item \textbf{Infrastructure Management:} For infrastructure management, these embeddings are used to monitor and predict changes in utility networks, such as water or electricity grids. They help in optimizing maintenance schedules and detecting faults and resource allocation \cite{wang2023scalableresourcemanagementdynamic}.
\end{itemize}

Overall, dynamic graph embeddings provide valuable insights and predictions by capturing the temporal evolution of graph structures and relationships, enabling accurate modeling and understanding of dynamic systems in various domains.

\section{Graph Embedding Stability}

Node embeddings have emerged as a powerful technique for representing nodes in complex networks. They capture the structural and semantic information of nodes in a low-dimensional vector space, enabling various downstream tasks such as link prediction, node classification, and recommendation systems. However, the rapid evolution of these methods has also introduced new challenges, particularly in the realm of temporal stability for dynamic graph representations.

This thesis focuses on addressing the stability issues in dynamic graph representation learning, a critical aspect that has emerged as graph embedding methods have been applied to time-varying networks. By investigating temporal consistency and structural fidelity, we aim to enhance the reliability of evolving network models, contributing to the ongoing advancement of graph representation techniques.

A key trait of a reliable graph representation method is that
it should exhibit embedding stability, which can be defined for a static graph through different runs or parameter initializations, etc or for a dynamic graph to analyze embedding stability through time. Understanding the stability of node representation models is crucial for designing effective node embedding methods and maximizing their utility in network analysis.

\subsection{Graph Embedding Stability in Static Settings}

Works in this section examine what we term "node embedding stability" in a static context, focusing on consistency across different runs or configurations of node embedding models. While these studies use similar terminology to our approach, we actually consider them under Sensitivity and Robustness Analysis as they analyze how sensitive the embeddings are to changes in model parameters or initialization, and how robust they are across different runs. This distinction is important as it differentiates between static and dynamic stability concepts. We will review key works in this area, highlighting their contributions to understanding embedding stability in static graph settings.

In \cite{9076342}, authors define the stability of network embedding as the invariance of the nearest neighbors of nodes in different embedding spaces that have been yielded from different runs. They investigate the potential influence factors on stability, specifically examining network structure properties and algorithm properties. This work sheds light on the fact that instability has remarkable impacts on the performance of network embedding in downstream applications. 

In \cite{schumacher2020effects}, authors define embedding stability informally as the difference between two embeddings due to randomness in the embedding algorithm and introduce three similarity metrics to measure geometric stability and one metric to assess method’s stability under randomness with respect to their performance in downstream tasks such as node classification and link prediction. They look into the influence of node centrality, network size and density on the geometric stability of node embeddings. 
Also in \cite{klabunde2022prediction}, authors assessed the instability of graph neural network predictions with respect to random initialization, model architecture, data, and training setup. Their experiments show that multiple instantiations of popular GNN models trained on the same data with the same model hyperparameters result in almost identical aggregated performance, but display substantial disagreement in the predictions for individual nodes. Authors show that instability of GNNs is reflected in their internal representations and maximizing model performance almost always implicitly minimizes prediction instability.

In \cite{agarwal2021unified} and \cite{9679093}, representation stability is measured by the instability score, defined as the percentage of test nodes for which predicted label changes when random noise is added to node attributes. In \cite{agarwal2021unified}, authors show that representations generated by their method, NIFTY, are stable, where an encoder function is considered stable according to the notion of Lipschitz continuity. 

In \cite{alsayouri2018recs}, authors evaluate their proposed model, RECS, in terms of representation learning stability, through two measures. (1) Representation Stability, by verifying the similarity of the learned vectors across different independent runs of the algorithms, and (2) Performance
Stability, where they use embeddings from different runs in a classification task
and we measure the variation in the classification performance. Ideally, a robust
embedding should satisfy both criteria.

In \cite{shi2023geometric}, authors consider a graph $G = (V, E)$ with $|V|$ nodes and $|E|$ relationships. Each node $v_i$ in the graph has a final embedding $z_i \in \mathbb{R}^{1 \times d}$. They denote the embeddings of all nodes in the graph as $Z \in \mathbb{R}^{|V| \times d}$. They denote the embeddings from configuration $k$ as $Z^k$, let $N$ denote the overall number of configurations. The authors present the Graph Gram Index (GGI), described in Algorithm~\ref{alg:ggi}, as a metric for evaluating the stability of embeddings across various configurations. 

\begin{algorithm}
\caption{Graph Gram Index (GGI) \cite{shi2023geometric}}
\begin{algorithmic}[1]
\Procedure{GGI}{$Z_1, Z_2, \ldots, Z_N$}
\State \textbf{Input:} $Z_1, \ldots, Z_N$
\State \textbf{Output:} Stability index $s \in [0, 1]$
\For{$l \in [1, \ldots, N]$}
    \State Center and normalize $Z_l$
    \State $S_l = A \circ (Z_l Z_l^T)$
    \State $s_l = \frac{1}{2|E|} \sum_{i,j \in |V| \times |V|} S_l[i, j]$
\EndFor
\State $s = \text{std}(s_l : l \in [1, \ldots, N])$
\EndProcedure
\end{algorithmic}
\label{alg:ggi}
\end{algorithm}

The Gram matrix $Z^l (Z^l)^T$ is used to capture the covariance information between embeddings $z_i^l$ and $z_j^l$, avoiding the need for alignment. Intuitively, this is achieved because they compare within one set of embeddings to generate a
structure summary, and compare the summaries across configurations. By applying a Hadamard product between the adjacency matrix and the Gram matrix, they focus on node pairs that are actual edges in the graph, ensuring that a stable model maintains the similarity of embeddings for these node pairs across different embedding spaces, regardless of any equivariant transformations.

In the context of word embeddings, there are related works which also look into the stability of embeddings. In particular, Borah et al. \cite{10.1145/3465336.3475098} investigate the stability of word embedding methods and its implications for natural language processing tasks. The study focuses on three word embedding methods: Word2Vec, GloVe, and fastText, evaluating their stability across multiple runs using intrinsic evaluation based on word similarity. The authors define word embedding stability in terms of the consistency of nearest neighbors for a given word across different embedding spaces generated from multiple runs with varying training parameters. Specifically, they measure stability as the number of shared nearest neighbors that a word's embeddings have in two different spaces, averaged over the entire vocabulary, as below:

\begin{align}
\text{Stability}(A) &= \frac{1}{\text{VocabSize}} \sum_{i=1}^{\text{VocabSize}} \text{stability}(w_i) \\
\text{where stability}(w) &= \frac{|KNN_{E_1}(w) \cap KNN_{E_2}(w)|}{k}
\end{align}

Here, $KNN_E(w)$ represents the set of top $k$ nearest neighbors of word w in the embedding space E. The research examines four diverse datasets and explores the effect of word embedding stability on downstream tasks such as clustering, POS tagging, and fairness evaluation. Their findings indicate that among the three word embedding methods, fastText demonstrates the highest stability, followed by GloVe and Word2Vec. However, they conclude that word embedding methods are not completely stable. The study highlights the importance of considering word embedding stability in NLP applications and provides insights into the factors affecting embedding consistency across different runs.

Similarly, in \cite{antoniak-mimno-2018-evaluating}, authors work on identifying the factors that create instability and measure statistical confidence in the cosine similarities between word embeddings trained on small, specific corpora. More specifically, the Jaccard coefficient for the n most similar words is used to measure stability in this work. This measurement for stability was also used in \cite{hellrich-etal-2019-influence} to compare word embedding algorithms and evaluate their stability and accuracy. Also, \cite{a58b9be8ce8a4c09a4d6f948ff3d5a53} and \cite{gong-etal-2020-enriching} are additional studies that evaluate word embeddings with a similar concept of temporal stability as previously discussed works in this area.

\subsection{Graph Embedding Stability in Temporal Settings}

Temporal node embedding methods are designed to capture the dynamic nature of evolving networks by representing nodes in a way that reflects their changing relationships over time. These methods typically process sequences of network snapshots or continuous event streams to model how node representations evolve. By incorporating temporal information, these embeddings offer a more sophisticated understanding of network dynamics, which can lead to more accurate predictions and analyses in various applications \cite{10.1016/j.neucom.2021.03.138}.

A critical aspect of dynamic node embedding methods is their temporal stability \cite{davis2023simplepowerfulframeworkstable, 10.5555/3540261.3541038}. Temporal stability refers to the consistency between changes in node embeddings and changes in the structure of the dynamic graph across different time steps. In temporal networks, embedding stability is crucial for ensuring robustness and reliability in downstream tasks \cite{modell2023intensity}. Significant fluctuations in node embeddings between time periods without underlying cause in the dynamics of graph data can impair the performance and generalizability of predictive models and analytical algorithms.

Building on this understanding of temporal stability, our work introduces two key concepts - temporal consistency and structural fidelity - to provide a more comprehensive framework for assessing and improving the reliability of dynamic graph representations. 

Temporal consistency measures how well changes in a node's embedding reflect structural changes in the dynamic graph over time. Structural fidelity ensures that the proximity of node embeddings accurately represents their structural relationships at each time step. By examining these aspects, we aim to provide a comprehensive framework for assessing and improving the reliability of dynamic graph representations, addressing crucial challenges in capturing evolving network structures.

In \cite{goyal2018dyngem}, a dynamic graph representation learning method is proposed and authors introduced the concept of stability in dynamic graph embedding in this work for the first time. In this work, first they define relative stability as 

\begin{equation}
    S(F; t) = \frac{\frac{||F_{t+1}(V_t)-F_{t}(V_t)||_F}{||F_{t}(V_t)||_F}}{\frac{||S_{t+1}(V_t)-S_{t}(V_t)||_F}{||S_{t}(V_t)||_F}}
    \label{eq:goyal18}
\end{equation}

where $V_t$ is the node set at time $t$, $S_t(\tilde{V})$ is the weighted adjacency matrix of the induced subgraph of node set $\tilde{V} \subseteq V_t$ and $F_t(\tilde{V}) \in \mathbb{R}^{|\tilde{V}| \times d}$ is the embedding of all nodes in $\tilde{V} \subseteq V_t$ of snapshot $t$.
 
Then the stability constant is defined as 
\begin{equation}
    K_S(F) = \underset{\tau, \tau^{'}}{\max} |S(F, \tau) - S(F, \tau^{'})|
    \label{eq:goyal218}
\end{equation} 

In this work, a dynamic embedding $F$ is considered stable as long as it has a small stability constant. The smaller the $K_S(F)$ is,
the more stable the embedding $F$ is. 

Through experiments they demonstrate the stability of their technique across time and show that their method maintains its competitiveness on evaluation tasks like graph reconstruction, link prediction and visualization.

In \cite{gürsoy2021alignment}, authors explore the embedding alignment and its parts, propose metrics to measure alignment and stability, and show their practicality through synthetic experiments. They define the stability error simply as 
\begin{equation}\frac{1}{|V|}{\sum_{i=1}^{N} \frac{||n_{i}^{t+1} - n_i^{t}||}{||n_i^{t}|| + ||n_i^{t+1}||}}
\label{eq:gursoy21}
\end{equation}
where $n_i^t$ is the embedding of node i at time t. This definition gives information about the temporal continuity of the embeddings. In their findings, they do not observe very meaningful differences between static and dynamic methods with respect to the defined stability error. They add that still, dynamic methods usually produce smaller stability errors possibly due to the information sharing between timesteps.

In \cite{DBLP:journals/snam/GursoyB22}, authors identify a set of classical migration laws and examine them via various methods such as temporal stability analysis. They use the stability error defined in \cite{gürsoy2021alignment}, Equation \ref{eq:gursoy21},  to reflect the extent of structural change in dynamic embeddings in this network.

In \cite{10.5555/3540261.3541038}, the authors introduce the concepts of cross-sectional and longitudinal stability and demonstrate that UASE \cite{Jones2020TheMR} meets both stability criteria. They provide a motivating example to illustrate that these two forms of stability are beneficial for the spatio-temporal representation of nodes. First, cross-sectional stability, which involves attributing the same position, up to a noise, to nodes that behave similarly at a given time. Second, longitudinal stability, which entails assigning a constant position, up to noise, to an individual node that demonstrates similar behavior across time. They find that existing dynamic latent position models [\cite{pensky2017spectral},\cite{10.1145/3269206.3271740}, \cite{7511675}] tend to trade one type of these stabilities off against the other. 

In \cite{davis2023simplepowerfulframeworkstable}, authors propose that a wide class of static network embedding methods can be used to produce interpretable and powerful dynamic network embeddings when they are applied to the dilated unfolded adjacency matrix. They define two forms of stability: Spatial Stability and Temporal Stability. A dynamic embedding is spatially stable if, for nodes with identical properties at the same time step, their embeddings are exchangeable:

\[
P(Y^{(t)}_i = v_1, Y^{(t)}_j = v_2) = P(Y^{(t)}_i = v_2, Y^{(t)}_j = v_1)
\]

A dynamic embedding is temporally stable if, for nodes with identical properties at different time steps, their embeddings are exchangeable:

\[
P(Y^{(u)}_i = v_1, Y^{(t)}_i = v_2) = P(Y^{(u)}_i = v_2, Y^{(t)}_i = v_1)
\]

They provide a theoretical guarantee that these
unfolded methods will produce stable embeddings. Furthermore, they considered a range of trivial simulated networks with planted spatial or temporal changes and introduced a hypothesis test which can be used to evaluate how
well dynamic embedding methods encode structure. Even in the simplest possible case, unstable dynamic embedding methods are at best conservative, but more often encode incorrect structure. 
They finally demonstrated the practical advantage that
unfolded methods possess over unstable methods by applying them to a pandemic-disrupted
dynamic flight network. Here, only the unfolded methods could encode the periodicity and
abrupt changes present in the network.

In \cite{modell2023intensity}, the authors introduce two notions for ensuring the stability of node embeddings: structural coherence and temporal coherence. Structural coherence is defined as the condition where, if two nodes exhibit  statistically indistinguishable behaviour at a given point in time, their representations at that time are close. That is, if \( \Lambda_i(t) = \Lambda_j(t) \), then \( \hat{X}_{i}(t) \approx \hat{X}_{j}(t) \), where $\Lambda_i(t)$ corresponds to node $i$'s behaviour at time $t$ and $\hat{X}_{i}(t)$ shows node $i$'s embedding at time $t$.
Temporal coherence is defined as the condition where, if a node exhibits  statistically indistinguishable behaviour at two different points in time, its representations at those times are close. That is, if \( \Lambda_i(s) = \Lambda_i(t) \), then \( \hat{X}_{i}(s) \approx \hat{X}_{i}(t) \). In this work authors present a representation learning framework, Intensity Profile Projection,
for continuous-time dynamic network data which satisfies both these conditions. 

The use of temporal stability in anomaly detection for dynamic graphs offers advantages such as capturing evolving patterns and distinguishing between normal network evolution and unusual events. However, challenges include sensitivity to noise and computational complexity for large-scale graphs. DynGEM \cite{goyal2018dyngem}, for instance, uses Euclidean distance between graph embeddings of consecutive timesteps for event detection, demonstrating one application of this concept.

%% file: Chapters/methodology.tex
\chapter{Methodology}
\label{chap:methodology}

In this work, we introduce the notion of representation integrity, or truthfulness, in dynamic graph learning, and  examine how well dynamic graph learning models align with actual changes in dynamic graphs. Our methodology focuses on an evaluation framework that assesses and compares models based on their temporal integrity. This refers to how accurately the models' learned representations capture the evolving nature of dynamic graphs over time.

First we design a metric to quantify temporal integrity in dynamic graphs which is comprised of three parts: a graph measure, a representation measure, and an alignment kernel. We define a set of integrity metrics in this space and compare and validate them through a set of desired properties, as well as desired behaviour in synthetic scenarios. After designing an integrity metric that satisfies these, we use this selected metric to compare dynamic graph models in terms of their temporal integrity.  This  evaluation framework built upon our validated indexes can provide a complementary evaluation tool for temporal graph learning models, to help address some of the evaluation challenges mentioned in recent works \cite{Poursafaei2022TowardsBE, bechlerspeicher2025positiongraphlearninglose}.
Our representation integrity metric can serve as a guiding principle in designing dynamic graph learning models.


\section{Problem Definition}

Let $\mathcal{G} = \{G_{t}\}_{t=1}^{T}$ be a \textit{discrete-time dynamic graph}, where each snapshot $G_t = (V, E_t)$ consists of a node set $V$ (fixed for simplicity) and a time-varying edge set $E_t \subseteq V \times V$. A dynamic graph embedding model $f$ maps each snapshot $G_t$ to a set of node representations:
\[
    R_{t} = f(G_{t}) \in \mathbb{R}^{|V| \times d},
\]
where the $d$-dimensional row vector $\mathbf{r}_{t}(v)$ represents node $v \in V$ at time $t$.

While static graph metrics (e.g., reconstruction loss, neighborhood preservation) are often used to assess embedding quality at individual time steps-what we term ``Static Integrity''-this work focuses exclusively on evaluating \textbf{Temporal Integrity}: the degree to which changes in the learned representations reflect meaningful changes in the underlying graph over time.
A representation is considered to have temporal integrity if it evolves in alignment with graph's dynamics.

\section{Metric Design}

To assess Temporal Integrity in dynamic graph learning models, we propose a general metric design space, consisting of three modular components:
\begin{enumerate}
    \item A \textbf{Graph Change Measure} $\Delta^G_t$, quantifying graph dynamics changes from $G_t$ to $G_{t+1}$.
    \item A \textbf{Representation Change Measure} $\Delta^R_t$, capturing how node embeddings change from $R_t$ to $R_{t+1}$.
    \item An \textbf{Alignment Kernel} $I(\cdot, \cdot)$, measuring the similarity or consistency between $\Delta^G_t$ and $\Delta^R_t$ through time.
\end{enumerate}

For each component we consider several candidate measures commonly used in the literature \cite{Tantardini2019ComparingMF}; their definitions are given in Section~\ref{sec:indexcomponents}.

For each time step $t$, we compute:
\[
  \text{TemporalIntegrity}(t) = I(\Delta^G_t, \Delta^R_t),
\]
and define \textbf{Total Integrity Index (TII)} as the average over time:
\[
  \text{TII} = \frac{1}{T-1} \sum_{t=1}^{T-1} \text{TemporalIntegrity}(t).
\]

\section{Metric Evaluation}
We evaluate a set of indexes composed in the metric design space based on a set of desired properties, and validate them empirically on synthetic data to identify the most effective Integrity Indexes. We then use the validated indexes to compare dynamic graph representation learning models based on how well their embeddings reflect the actual evolution of the graph. Through this formulation, we establish \textbf{Temporal Integrity}-as captured by our composable, interpretable metric design space-as a principled criterion for evaluating the integrity of dynamic graph representations.

\subsection{Properties}

For this section, we assume all the graph change and representation change measures can be bounded to $[0,1]$, that is $\Delta^{G}_t := \Delta(G_t, G_{t+1}) \in [0, 1]$ and $\Delta^{R}_t := \Delta(R_t, R_{t+1}) \in [0, 1]$. With this assumption, 
we propose the following properties that a well-designed integrity index $I_t$ should satisfy:

\begin{itemize}

\item \textbf{Property 1 (Boundedness).}
The integrity score is confined to $[0,1]$:
\[
    0 \le I_t \le 1 \qquad \forall t .
\]

This property guarantees interpretability and comparability across graphs and models.

\item \textbf{Property 2 (Optimal Alignment)}. 
The index reaches its maximum when the representation respond perfectly to the true change.
\[ \Delta^{G}_t = \Delta^{R}_t \Rightarrow I_t = 1 \]

\item \textbf{Property 3 (Upper Anchor).}
For a representation that is theoretically stable,
$\, I_{\text{t}}\,$ must be close to 1:
\[
    I_t(\text{Stable Model}) \sim 1 .
\]

If method is not stable, index should be significantly lower.

\end{itemize}

\vspace{0.3cm}
\noindent
\subsection{Synthetic Scenarios}
\label{sec:scenarios}

We will empirically test these properties using synthetic dynamic graphs generated with Stochastic Block Models (SBMs), under the following scenarios:

\newcommand{\Nochange}{Constant Graphs\xspace}
\newcommand{\Merge}{Gradual Merge\xspace}
\newcommand{\Move}{Abrupt Move\xspace}
\newcommand{\Periodic}{Periodic Transitions\xspace}

\begin{itemize}
    \item \textbf{\Merge}: Two equally sized communities start almost disconnected; the probability of cross-community edges rises linearly from
    a low to high probability over $T$ snapshots.

    \emph{Motivation:} tests a model’s ability to track
        \textit{smooth, long-horizon drift}; representations must evolve continuously without over-reacting to small incremental edits.

    \item \textbf{\Move}:  
    Three equally sized communities start with high  within-community edge probability and only weak ties to other communities (low cross-community edge probability). Throughout the first half of the sequence of length $T$, the connection probabilities remain as described; at the midpoint, the edge connection probabilities change, such that the nodes in community $1$ migrate to community $0$ and adopt the edge connection probabilities of community $0$. This new behaviour remains fixed for the rest of the experiment.  
    
    \emph{Motivation:} probes \textit{reactivity to sudden,
        localized change}; the model must detect and accommodate a sharp structural jump rather than smoothing it away.

    \item \textbf{\Periodic}:  
      Two equally sized communities start with high probability of within-community edges and low cross-community edges. Their interaction pattern follows a smooth cyclic rhythm.  Within-community
      edges become more (or less) likely according to a sinusoid, rising from a low to high probability and back again every $\frac{T}{5}$ snapshots; between-community
      probabilities vary inversely. The graph therefore alternates between phases of clear modular structure and phases of near mixing, in a regular cycle.

      \emph{Motivation:} evaluates whether the model retains
        \textit{temporal memory} and can represent patterns that repeat
        over long intervals, a capability needed for forecasting
        seasonality or scheduling effects in real networks.

     
\end{itemize}

\section{Index Components}
\label{sec:indexcomponents}

\subsection{Graph–Change Measures}
\label{sec:deltaG}

To quantify how much the graph evolves from $G_t$ to $G_{t+1}$ we employ three distance measures. Each one is normalized to the unit interval \([0,1]\) so that integrity scores remain comparable across datasets and scenarios. We include three distances that offer diverse perspectives: adjacency-based (Euclidean), affinity-based (DeltaCon), and spectral.

\subsubsection{Scaled Euclidean Adjacency Distance}
Let \(A_t\) and \(A_{t+1}\) be the adjacency matrices on the common node
set of size \(n\).  We use the Frobenius norm,
\[
d_{\mathrm{EUC}}(G_t,G_{t+1})
  = \frac{\lVert A_t - A_{t+1}\rVert_F}{\sqrt{n(n-1)}},
\]
where the denominator is the maximum possible difference (an empty
versus a complete graph), yielding a value in \([0,1]\).  The measure
captures aggregate edge-level changes and is quick to compute.


\subsubsection{DeltaCon Distance}
DeltaCon similarity~\cite{Faloutsos2013DELTACONAP} quantifies the resemblance between two graphs
\(G_{1}=(V,E_{1})\) and \(G_{2}=(V,E_{2})\) sharing the same node set.

Let \(A_{i}\in\{0,1\}^{n\times n}\) be the (possibly weighted) adjacency matrix of \(G_{i}\) and
\(D_{i}=\operatorname{diag}(A_{i}\mathbf 1)\) its degree matrix.
For a small diffusion parameter \(\epsilon>0\) we form the node-affinity
\emph{Fast Belief Propagation} matrix
\[
  S_{i} \;=\; \bigl(I + \epsilon^{2}D_{i} - \epsilon A_{i}\bigr)^{-1},
  \qquad i\in\{1,2\},
\]
whose entry \(S_{i}(u,v)\) reflects the ease with which information can flow between nodes \(u\) and \(v\) in \(G_{i}\).  It can be shown that sij depends on all the r-paths connecting (i, j) ~\cite{Faloutsos2013DELTACONAP}.

DeltaCon similarity is then defined as
\[
  \operatorname{DeltaCon}_{\mathrm{sim}}\!\bigl(G_{1},G_{2}\bigr)
  \;=\;
  \frac{1}{
    1 + \displaystyle\sqrt{\sum_{u,v\in V}
      \!\bigl(\sqrt{S_{1}(u,v)} - \sqrt{S_{2}(u,v)}\bigr)^{2}}
  }
\]
which attains \(1\) when the graphs are identical and decreases smoothly toward \(0\) as they diverge.

We convert this similarity into a \textbf{graph-change distance} by taking
\[
  \operatorname{d}_{\mathrm{DeltaCon}}\!\bigl(G_{t},G_{t+1}\bigr)
  \;=\;
  1 - \operatorname{DeltaCon}_{\mathrm{sim}}\!\bigl(G_{t},G_{t+1}\bigr)
  \;\in\; [0,1],
\]
so that larger values directly indicate greater change between snapshots.

It can be shown \cite{Faloutsos2013DELTACONAP, 10.1145/2824443} that DeltaCon satisfies several desirable properties: it penalizes changes that disconnect a graph more heavily; in weighted graphs, removing higher‑weight edges increases the distance more; alterations in sparse graphs have greater impact than identical changes in denser graphs; and random modifications yield smaller distance increases than targeted ones.\\

\subsubsection{Laplacian Spectral Distance}
Let \(L_t\) and \(L_{t+1}\) be the Laplacians of $A_t$ and $A_{t+1}$ and
\(\lambda^{(t)}_1\le\cdots\le\lambda^{(t)}_n\) their eigenvalues.  We
define
\[
d_{\mathrm{SPEC}}(G_t,G_{t+1})
  = \frac{\bigl\lVert
      \boldsymbol\lambda^{(t)}-
      \boldsymbol\lambda^{(t+1)}
    \bigr\rVert_2}{\,n\sqrt{\,n-1\,}},
\]
where \(\boldsymbol\lambda^{(t)}\) is the sorted eigenvalue vector.
Norm-2 difference of \(\boldsymbol\lambda^{(t)}\) and \(\boldsymbol\lambda^{(t+1)}\), is divided by the maximum possible value $n\sqrt{n-1}$, which is between an empty and a complete graph, to normalize the measure in to $[0, 1]$ internal. This measure captures differences in global connectivity patterns and community structure.

\subsubsection{Spectral Distance (first–\(k\) eigenvalues)}
\label{sec:spectral-distance}

Let \( \widetilde L_t \) and \( \widetilde L_{t+1} \) denote the normalized Laplacians of \(G_t\) and \(G_{t+1}\), whose spectra lie in
\([0,2]\).
Denote by
\(0=\lambda^{(t)}_{0}<\lambda^{(t)}_{1}\le\cdots\le\lambda^{(t)}_{k}\)
and
\(0=\lambda^{(t+1)}_{0}<\lambda^{(t+1)}_{1}\le\cdots\le
  \lambda^{(t+1)}_{k}\)
the first \(k\) \emph{non-zero} eigenvalues of \( \widetilde L_t \) and \( \widetilde L_{t+1} \).

We measure spectral change as the Euclidean distance between these two eigenvalue vectors,
normalised by the largest possible distance, \(2\sqrt{k}\):

\[
d_{\mathrm{SPEC}}(G_t,G_{t+1})
  \;=\;
  \frac{\displaystyle
        \bigl\lVert
          (\lambda^{(t)}_{1},\dots,\lambda^{(t)}_{k})
          -
          (\lambda^{(t+1)}_{1},\dots,\lambda^{(t+1)}_{k})
        \bigr\rVert_{2}
       }
       {\sqrt{k}\,\,\cdot 2}\;\in[0,1].
\]

Using only the lowest \(k\) non-zero modes focuses the distance on
\emph{global} structural features-such as coarse community structure and
overall connectivity.

\subsection{Representation Change Measures}

To quantify how much a model’s node embeddings have changed from snapshot $t$ to $t+1$, we employ two scale‑free representation change measures.

In our experiments, we also implement a variant where we perform Procrustes alignment before calculating the representation change. This alignment eliminates discrepancies due to rotations and reflections in the representation space. We describe this variant in the Appendix~\ref{app:appendix}.

\paragraph{Row--Cosine Change.}

Every node representation vector is first $\ell_2$‑normalised
\[
\hat e_{t,i}= \frac{(E_tQ^{\star})_i}{\lVert(E_tQ^{\star})_i\rVert},\qquad
\hat e_{t+1,i}= \frac{(E_{t+1})_i}{\lVert(E_{t+1})_i\rVert},
\]
ensuring scale invariance; therefore it captures only directional (i.e., relational) changes, not arbitrary differences in vector magnitude.

For each node we compute the directional drift
\[
\Delta r^{cos}_{i} \;=\; 1 - \cos\!\bigl(\hat e_{t,i},\,\hat e_{t+1,i}\bigr)
           \;\in\;[0,1],
\]
where $0$ means the embedding direction is unchanged and $1$ means it has flipped to the opposite pole. The snapshot‑level representation change is defined as the bounded average:
\[
\Delta R^{cos}
   \;=\;
   \frac{1}{n}\sum_{i=1}^{n}\Delta r^{cos}_i
   \;\in\;[0,1].
\]

\paragraph{Row-Unit Euclidean Change.}
After projecting each node's representation vector onto the unit hypersphere, we measure the half-Euclidean distance:

\[
\Delta r^{euc}_{i} \;=\; \tfrac12 \,\bigl\lVert\hat e_{t,i} - \hat e_{t+1,i}\bigr\rVert
      \;\in [0,1].
\]

The snapshot‑level representation change is defined as the bounded average:
\[
\Delta R^{euc}
   \;=\;
   \frac{1}{n}\sum_{i=1}^{n}\Delta r^{euc}_i
   \;\in\;[0,1].
\]

\subsection{Alignment Kernels}

We use seven candidate alignment kernels to calculate the integrity index \( I_t \), which measures the agreement between graph changes and representation changes. Each of the alignment kernels considered is bounded within the range \([0,1]\) by design and reaches the theoretical extrema, as stipulated by Property 2 ($\Delta^{G}_t = \Delta^{R}_t \Rightarrow I_t = 1$).

\subsubsection{Symmetric Deviation Penalty}
The kernel
\[
I^{Sym}_t = 1 - |\Delta^R_t - \Delta^G_t|
\]
computes integrity as the complement of the absolute difference between representation change $\Delta^R$
and graph change $\Delta^G$. This linear penalty ensures simplicity and symmetry, treating overreaction and underreaction equivalently.

\subsubsection{Gaussian Similarity}
Defined as
\[
I^{Gauss}_t = \exp\left(-\frac{(\Delta^R_t - \Delta^G_t)^2}{2\sigma^2}\right)
\]

this kernel uses a Gaussian function to smoothly penalize misalignment. The sensitivity parameter $\sigma$ controls the rate of decay in integrity scores as $\Delta^G$ diverges from 
$\Delta^R$.






\subsubsection{KL-Divergence Inspired}

This kernel is based on the Kullback–Leibler (KL) divergence, a fundamental concept from information theory that quantifies how one probability distribution diverges from a reference distribution. The standard KL divergence for discrete distributions $P$ and $Q$ is defined as
\[
D_{\text{KL}}(P \parallel Q) = \sum_{x} P(x) \log\left(\frac{P(x)}{Q(x)}\right),
\]
and is commonly used to measure the information loss when $Q$ is used to approximate $P$.

In our context, we adapt this idea to compare the scalar change measures $\Delta^G_t$ and $\Delta^R_t$, which are interpreted as probabilities of change in a Bernoulli setting. The kernel is defined as
\[
I^{KL}_t = 1 - \left[ \Delta_t \log\left(\frac{\Delta_t}{\delta_t + \epsilon}\right) + (1 - \Delta_t) \log\left(\frac{1 - \Delta_t}{1 - \delta_t + \epsilon}\right) \right],
\]
where $\epsilon > 0$ is a small constant for numerical stability.

The formulation applies the KL divergence between two Bernoulli distributions, with $\Delta^G_t$ as the reference and $\Delta^R_t$ as the approximation. The result is then subtracted from $1$ to yield a similarity score, so that higher values of $I_t$ indicate better alignment between the representation and the graph change. The smoothing term $\epsilon$ ensures the kernel remains well-defined even when $\Delta^G_t$ is close to $0$ or $1$.

\subsubsection{Pearson Correlation}
A global linear association is captured by the Pearson correlation coefficient
$\rho_{\mathrm{pearson}}\!\in[-1,1]$:
\begin{equation}
  I^{\mathrm{pearson}}_{t}(\Delta^{G}_t,\Delta^{R}_t) \;= \tfrac12\bigl(\rho_{\mathrm{pearson}}(\Delta^G_t,\Delta^R_t)+1\bigr)
\end{equation}

\paragraph{Spearman Correlation}
Spearman’s correlation coefficient is a statistical measure of the strength of a monotonic relationship between paired data. Spearman correlation $\rho_{\mathrm{spearman}}\!\in[-1,1]$ works by calculating Pearson’s correlation on the ranked values of the data. 

\begin{equation}
  I^{\mathrm{spearman}}_t(\Delta^G_t,\Delta^R_t) \;=\;
    \tfrac12\bigl(\rho_{\mathrm{spearman}}(\Delta^G_t,\Delta^R_t)+1\bigr)
\end{equation}

\paragraph{Time Lagged Cross Correlation}
Representation models may respond to a structural shock after a short delay.  Let
$L\!\ge\!0$ be the maximum lag we are willing to tolerate.  We scan all shifts within $\pm L$ and keep the best Pearson correlation:
\begin{equation}
  I^{\mathrm{xcorr}}_t(\Delta^G_t,\Delta^R_t)
  \;=\;
  \tfrac12\!\Bigl(\max_{|\ell|\le L}\rho_{\mathrm{pearson}}\!\bigl(\Delta^G_{0:T-\ell},
                                               \Delta^R_{\ell:T}\bigr)
                  + 1\Bigr).
\end{equation}

In our experiments we set $L=3$, trading off
responsiveness against the risk of hiding genuinely mistimed embeddings.

\paragraph{Dynamic Time Warping}
Finally, Dynamic Time Warping (DTW) aligns the two trajectories under arbitrary monotone, contiguous warps.  Let $\operatorname{DTW}(\cdot,\cdot)$ denote the
classic DTW distance and $T$ the sequence length:
\begin{equation}
  I^{\mathrm{DTW}}_t(\Delta^G_t,\Delta^R_t)
  \;=\;
  \max\!\Bigl(0,\,
  1-\frac{ \operatorname{DTW}(\Delta^G_t,\Delta^R_t)}{T}\Bigr).
  \label{eq:k-dtw}
\end{equation}
Normalising by $T$ ensures scale invariance; truncating negative values at zero
interprets extremely large misalignments as complete loss of integrity.

\section{Representation Alignment}
In this section, we describe another approach for computing integrity scores that first applies Procrustes alignment to the node embeddings before integrity score computation. Specifically, to compare embeddings from snapshot $t$ to $t{+}1$, we first align the two embedding matrices $E_t,E_{t+1}\!\in\!\mathbb{R}^{n\times d}$ using the optimal orthogonal matrix, 
\[
Q^{\star}\;=\;\arg\min_{Q^{\mathsf T}Q = I}\,
                 \lVert E_t\,Q - E_{t+1}\rVert_F,
\]
obtained through singular-value decomposition of $E_t^{\!\mathsf T}E_{t+1}$. This \emph{Procrustes} alignment eliminates arbitrary rotation or reflection that occurs when models are retrained independently at each time step-variations that should \emph{not} count as semantic drift.

%% file: Chapters/experiments.tex
\chapter{Experiments}

The primary aim of our experimental study is twofold:
\begin{enumerate*}[label=\roman*)]
    \item To evaluate and select the most reliable integrity indexes for dynamic graph representation learning, using both theoretical and empirical criteria.

    \item To compare dynamic graph embedding models in terms of temporal integrity, as measured by the validated indices.
\end{enumerate*}

First, we validate the properties of each candidate index by examining its empirical behavior in each of the synthetic scenarios. To guide our analysis, we leverage UASE~\cite{Jones2020TheMR} and IPP~\cite{modell2023intensity} models as baselines for further analysis of the candidate indexes, since these models are mathematically proven to maintain integrity. \cite{10.5555/3540261.3541038}.
Based on this analysis, we select a validated index that meets all of our criteria. This allows for a straightforward comparison of temporal integrity of different learning-based models, which we report next. Additionally, we compute the correlation of the integrity scores of these models with their performance in link prediction, which is the common metrics used to evaluate dynamic graph models, to show how a more stable model also generally performs better on the downstream link prediction task.

\section{Experimental Setup}

\subsection{Models}

We evaluate several dynamic graph embedding models, including both learning-based and non-learning-based approaches. The primary models considered are:

\begin{itemize}
\item UASE (Unfolded Adjacency Spectral Embedding) \cite{Jones2020TheMR}: This method extends the random dot product graph model to handle dynamic multilayer networks by jointly embedding adjacency matrices across multiple time steps. It computes a singular value decomposition of the unfolded adjacency matrix to obtain node embeddings that are consistent over time and across similar nodes, making it particularly suitable for spatio-temporal analysis of evolving graphs \cite{Jones2020TheMR, 10.5555/3540261.3541038}.

\item IPP (Intensity Profile Projection) \cite{modell2023intensity}: IPP is a spectral representation learning framework for continuous-time dynamic network data. It works by first smoothing the observed event counts to get a pair-wise intensity function for every edge, then finding the best rank-d subspace that minimizes an intensity-reconstruction error, and finally projecting each node’s time-varying intensity profile onto that subspace to obtain its embedding.

\item GAT (Graph Attention Network) \cite{veličković2018graph}: This method utilizes the Graph Attention Network (GAT), as described in Section \ref{gat}.

\item AE (Autoencoder)\cite{goyal2018dyngem}: This method uses deep autoencoder to learn the representation of each node in the graph.  
\item dynAE (Dynamic Autoencoder) \cite{Goyal2018dyngraph2vecCN}: This method models the interconnection of nodes within and acrosstime using multiple fully connected layers. It extends Static AE for dynamic graphs. 
\item dynRNN (Dynamic Recurrent Neural Network)\cite{Goyal2018dyngraph2vecCN} : This method uses sparsely connected Long Short Term Memory(LSTM) networks to learn the embedding. 
\item dynAERNN (Dynamic Autoencoder Recurrent Neural Network) \cite{Goyal2018dyngraph2vecCN}: This method uses a fully connected encoder to initially ac-quire low dimensional hidden representation and feeds this representation into LSTMs to capture network dynamics. 
\end{itemize}

Each method generates node embeddings of dimension d = 64 for each time step. \footnote {We use the implementation of some dynamic graph models accessible in \href{https://github.com/palash1992/DynamicGEM}{DynamicGEM, A Library for Dynamic Graph Embedding Methods}.}

\subsection{Synthetic Datasets}
To systematically evaluate integrity indexes, we generate synthetic dynamic graphs using Stochastic Block Models (SBMs) under three controlled scenarios described in \ref{sec:scenarios}:

\begin{itemize}
  \item \textbf{Gradual Merge.}  
        Two communities begin nearly separate; the between-community link probability increases linearly from $0.1$ to $0.9$ over the $T=100$ snapshots, so the blocks gradually merge into a single mixed community.

  \item \textbf{Abrupt Move.}  
        Three equal communities evolve steadily with $0.9$ within-community edge probability and $0.05$ cross-community edge probabilities until the midpoint, when one community suddenly migrates to another community and adopts its connectivity pattern, triggering a sharp structural change.

  \item \textbf{Periodic Transitions.}  
        At each snapshot the within-community connection probability rises and falls in a smooth sine-wave cycle, while the between-community probability changes inversely. With a $20$-step cycle repeated over the $T=100$ timesteps, the graph returns to the same community pattern every $20$ steps, mimicking a regular seasonal change.
        
\end{itemize}

Each synthetic graph consists of $n=300$ nodes and approximately $30,000$ edges per timestep (exact number varies by scenario and SBM parameters). Edge probabilities and community sizes are chosen to ensure non-trivial but interpretable dynamics.

\begin{table}[h]
\center
\begin{tabular}{|c|c|c|c|>{\centering\arraybackslash}m{5cm}|}
\hline
Scenario & \# Nodes & \# Total Edges & T \\
\hline
Merge & 300 & 3,136,569 & 100  \\
\hline
Move & 300 & 1,701,030 & 100 \\
\hline
Periodic & 300 & 2,242,381 & 100 \\
\hline
\end{tabular}
\caption{Synthetic Dataset Statistics}
\end{table}

\section{Index Validation}

We start with 42 candidate integrity configurations and progressively narrow them down to a single index that fulfills all of the properties and expectations.

Combining the property analysis and the UASE or IPP-based empirical evaluation, we identify the subset of indexes that (a) satisfy all properties (including empirical support for property 3), and (b) most frequently recognize UASE or IPP as the best model across scenarios. These indexes are recommended for practical use in dynamic graph representation learning.

We compute the integrity score \(I_t = I\!\bigl(\Delta^G_t,\Delta^R_t\bigr)\) for each consecutive timestep where \(t=1,\dots ,T-1\) in all synthetic datasets (merge, move, periodic) and summarize it by the time–average.

\begin{equation}\label{eq:Ibar}
  \bar I
  ~=~
  \frac{1}{T-1}
  \sum_{t=1}^{T-1} I_t .
\end{equation}

We report the time-averaged integrity scores in tables \ref{tab:integrity-merge} for merge scenario, \ref{tab:integrity-move} for move scenario, and \ref{tab:integrity-periodic} for periodic scenario. For every considered configuration of the integrity index $(\Delta G,\Delta R,\text{kernel})$ in each row, the best, second-best and third-best model scores are highlighted in \textcolor{gold}{gold}, \textcolor{silver}{silver} and \textcolor{bronze}{bronze}, respectively.

\subsection{Results Per Each Scenario}


Each candidate integrity index is first assessed against property 3, defined in chapter~\ref{chap:methodology}, per each scenario.


An integrity index validates property 3 by assigning a high score to stable methods and demonstrating a sharp contrast between integrity scores of a stable model compared to an unstable baseline.

We use UASE \cite{Jones2020TheMR} and IPP \cite{modell2023intensity} graph methods as stable models in this experiment, as they are the only models to the best of our knowledge that are mathematically proven to be consistent both cross-sectionally (across nodes) and longitudinally (across time) \cite{10.5555/3540261.3541038, modell2023intensity}. Hence their integrity scores should be close to 1.
 
For the unstable model, a static auto-encoder is trained on the synthetic datasets, then the \(T=100\) snapshot blocks of its embedding are permuted at random. The per-snapshot quality is unchanged, but temporal alignment is destroyed, so its integrity score should be significantly lower than UASE or IPP’s.

\begin{table}[t] \centering \caption{Integrity scores in merge setting.} \begin{adjustbox}{max width=\linewidth, max height=0.9\textheight, center} \input{integrity_2merge100_noalign.tex} \end{adjustbox} \label{tab:integrity-merge} \end{table}

\begin{table}[t] \centering \caption{Integrity scores in move setting.} \begin{adjustbox}{max width=\linewidth, max height=0.9\textheight, center} \input{integrity_3move100_noalign.tex} \end{adjustbox} \label{tab:integrity-move} \end{table}

\begin{table}[t] \centering \caption{Integrity scores in periodic setting.} \begin{adjustbox}{max width=\linewidth, max height=0.9\textheight, center} \input{integrity_periodic100_noalign.tex} \end{adjustbox} \label{tab:integrity-periodic} \end{table}

\subsection{Aggregated Results Across Scenarios}

UASE \cite{Jones2020TheMR} and IPP \cite{modell2023intensity} are mathematically proven to be temporally coherent \cite{10.5555/3540261.3541038, modell2023intensity}. We use these models as reference to validate our integrity index: We record in how many of the three synthetic datasets (\emph{merge}, \emph{move}, \emph{periodic}) the index ranks either IPP or UASE as the top model \(\bigl(\text{Win}_{\text{IPP|UASE}}\in\{0,1,2,3\}\bigr)\). An index that consistently identifies IPP or UASE as the model with the highest integrity is considered more reliable in practice.

\noindent
Table~\ref{tab:gap-wins1} evaluates every integrity configuration on two axes:

\begin{itemize}
\item \textbf{Comparing integrity score gap between stable (UASE and IPP) and unstable models}
      \[
          \Delta \bar{I}_{\text{UASE}}
            = \bar I_{\text{UASE}} - \bar I_{\text{shuffled}},
          \qquad
          \Delta \bar{I}_{\text{IPP}}
            = \bar I_{\text{IPP}}  - \bar I_{\text{shuffled}} .
      \]
      A large positive value means the metric assigns a higher score to the theoretically stable representation (UASE or IPP) compared to the unstable baseline.

\item \textbf{Win\(_{\text{UASE}|\text{IPP}}\)}  
      -the number of data scenarios (merge, move, periodic) in which the same configuration also ranks either UASE or IPP as the best model (range 0–3).
\end{itemize}

\vspace{4pt}
\noindent
We consider a configuration to satisfy the revised properties when
\[
   \Delta \bar{I}_{\text{UASE}}\ge 0.25,
   \quad
   \Delta \bar{I}_{\text{IPP}} \ge 0.25,
   \quad
   \text{Win}_{\text{UASE}|\text{IPP}} = 3 .
\]

Under these criteria, the combinations of \textbf{Euclidean~$\Delta G$ + Euclidean or Cosine ~$\Delta R$ +
Pearson, Spearman, or XCorr Alignment kernel} pass all three thresholds and are therefore
appropriate choices of integrity index. As their performance are on par, we choose \textbf{Euclidean~$\Delta G$ + Euclidean ~$\Delta R$ +
Pearson} as the final integrity index for the remainder of this thesis. Table~\ref{tab:wins} presents the results for this index in both versions, where representations are either aligned before the integrity score computation or used without modification.

\input{wins.tex}



\section{Model Comparison Using Validated Index}

Using the validated integrity metric with composition \emph{Euclidean $\Delta$G + Euclidean $\Delta$R + Pearson}, we score every model on the three synthetic streams.
Table~\ref{tab:comparison} reports the time-averaged integrity
$\bar I$; within each row the best, second-best and third-best scores are
highlighted in \textcolor{gold}{gold}, \textcolor{silver}{silver} and
\textcolor{bronze}{bronze} respectively.

\subsection{Results Across Scenarios}

\begin{itemize}
    \item \textbf{Gradual–Merge.}  
          UASE attains the highest integrity
          ($\bar I=0.996$), confirming its role as a stability oracle;  
          IPP follows with $0.872$.  
          Among neural models, the Graph Attention Network (GAT) outperforms the
          temporal models, indicating that slow, smooth drift can be captured without temporal memory.
    \item \textbf{Abrupt–Move.}  
          UASE leads ($0.990$) and UASE comes a close second ($0.966$), showing
          they both react well to a sudden, community-level shift.  
          The simple Auto Encoder based model (AE) ($0.552$) is the best of the learning-based models, narrowly edging the dynamic variants. Here the models that try to learn temporal dynamics-DynRNN, DynAE, DynAERNN, StConv-do not yet improve on a static auto-encoder, implying they struggle to track a sudden migration.
    \item \textbf{Periodic–Transitions.}  
          IPP and UASE again dominate ($0.999$ and $0.998$).  
          yet now a dynamic architecture finally stands out: StConv reaches 0.727, outscoring every other learned method by a wide margin and claiming the bronze rank overall. The temporal convolution evidently helps when the underlying graph truly follows a rhythmic pattern, whereas static models still plateau near 0.5.
\end{itemize}

\input{comparison0.tex}

\paragraph{Integrity versus downstream link prediction.}
To assess whether higher integrity aligns with better task performance we use the Spearman rank correlation, which captures monotone rather than strictly linear associations.  
For each scenario we correlate the eight integrity scores with the eight one-step link-prediction AUCs of the same models:
\[
\rho \bigl(\bar I,\mathrm{AUC}\bigr) \;=\;
\begin{cases}
0.786 & \text{Gradual-Merge}\\
0.216 & \text{Abrupt-Move}\\
0.671 & \text{Periodic}\\
\end{cases}
\qquad
\text{mean } \rho = 0.596 .
\]

The rank correlation remains strong for the smooth \emph{Merge} stream and
moderate for the \emph{Periodic} stream, but drops to a weak value in the
\emph{Move} scenario, where AUC appears less sensitive to integrity
differences.  On average (\(\rho\!\approx\!0.596\)) the monotone relation
persists: models that preserve temporal consistency tend-though not
uniformly-to achieve higher link-prediction accuracy.

\subsection{Results in Real Dataset}

In this section we compare the integrity scores of graph models trained on a real dataset, using the validated integrity index. For this experiment we use CanParl \cite{Poursafaei2022TowardsBE} Dataset, which is a dynamic political network that tracks the interactions of Canadian Members of Parliament (MPs) from 2006 to 2019. Each node represents an MP, and an edge connects them if both voted "yes" on a bill.

The results are shown in table~\ref{tab:integrity_real}. The scores show a clear hierarchy: STConv dominates, achieving a near-perfect integrity score (0 .980), well ahead of every other method. The only models in its vicinity are the static auto-encoder (AE, 0 .671) and the non-learning baseline IPP (0 .647), which take silver and bronze, respectively. All remaining dynamic variants (DynAE, DynRNN, DynAERNN) cluster in the mid-0.5 range, indicating that the extra temporal complexity offered no advantage here, while the graph-attention model (GAT, 0 .404) trails far behind. In short, a simple spatio-temporal convolution captures the scenario’s structure best, static AE delivers a solid but distant second, and most dynamic or attention-based models contribute little beyond baseline performance.


\begin{table}[htbp]
  \centering
  \setlength{\tabcolsep}{6pt}
  \renewcommand{\arraystretch}{1.2}

  \caption{Integrity scores in CanParl Dataset.}
  \label{tab:integrity_real}

  \begin{tabular}{|l|c|c|c|c|c|c|c|c|}
    \hline
    & \textbf{IPP} & \textbf{UASE} & \textbf{GAT} & \textbf{AE}
    & \textbf{DynAE} & \textbf{DynRNN} & \textbf{DynAERNN} & \textbf{STConv} \\
    \hline
    \textbf{Integrity score}
      & \cellcolor{bronze}0.647
      & 0.616
      & 0.404
      & \cellcolor{silver}0.671
      & 0.617
      & 0.552
      & 0.470
      & \cellcolor{gold}0.980 \\
    \hline
  \end{tabular}
\end{table}

%% file: integrity_2merge100_noalign.tex
\begin{tabular}{lllrrrrrrrrr}
\toprule
\multicolumn{3}{c}{}%
& \multicolumn{2}{c}{\textbf{Non-learning-based}}%
& \multicolumn{3}{c}{\textbf{Static}}%
& \multicolumn{4}{c}{\textbf{Dynamic}}\\
\cmidrule(lr){4-5}\cmidrule(lr){6-8}\cmidrule(lr){9-12}
$\Delta G$ & $\Delta R$ & kernel & IPP & UASE & GAT & AE & AE-Shuffled & DynAE & DynRNN & DynAERNN & STConv \\
\midrule
DeltaCon & Cosine & Sym & 0.288 & 0.814 & 0.648 & \cellcolor{bronze}0.860 & \cellcolor{silver}0.870 & \cellcolor{gold}0.885 & 0.783 & 0.764 & 0.234 \\
DeltaCon & Cosine & Gauss & 0.0 & 0.496 & 0.278 & \cellcolor{bronze}0.633 & \cellcolor{silver}0.668 & \cellcolor{gold}0.718 & 0.429 & 0.385 & 0.0 \\
DeltaCon & Cosine & KL & 0.0 & 0.912 & 0.488 & \cellcolor{bronze}0.939 & \cellcolor{silver}0.948 & \cellcolor{gold}0.955 & 0.449 & 0.416 & 0.002 \\
DeltaCon & Cosine & Pearson & \cellcolor{gold}0.795 & \cellcolor{silver}0.630 & 0.575 & 0.345 & 0.392 & 0.543 & 0.533 & 0.406 & \cellcolor{bronze}0.601 \\
DeltaCon & Cosine & Spearman & \cellcolor{silver}0.739 & \cellcolor{bronze}0.580 & 0.565 & 0.327 & 0.38 & 0.542 & 0.518 & 0.417 & \cellcolor{gold}0.748 \\
DeltaCon & Cosine & Xcorr & \cellcolor{gold}0.806 & \cellcolor{bronze}0.671 & 0.619 & 0.381 & 0.463 & 0.543 & 0.558 & 0.538 & \cellcolor{silver}0.756 \\
DeltaCon & Cosine & DTW & 0.288 & 0.826 & 0.67 & \cellcolor{bronze}0.879 & \cellcolor{silver}0.905 & \cellcolor{gold}0.922 & 0.817 & 0.802 & 0.234 \\
DeltaCon & Euclidean & Sym & 0.4 & 0.767 & \cellcolor{bronze}0.817 & 0.797 & 0.799 & 0.804 & \cellcolor{gold}0.918 & \cellcolor{silver}0.918 & 0.241 \\
DeltaCon & Euclidean & Gauss & 0.001 & 0.324 & \cellcolor{bronze}0.539 & 0.423 & 0.429 & 0.444 & \cellcolor{silver}0.829 & \cellcolor{gold}0.836 & 0.0 \\
DeltaCon & Euclidean & KL & 0.136 & 0.876 & 0.884 & 0.898 & 0.901 & \cellcolor{bronze}0.907 & \cellcolor{silver}0.976 & \cellcolor{gold}0.977 & 0.0 \\
DeltaCon & Euclidean & Pearson & \cellcolor{gold}0.799 & \cellcolor{silver}0.631 & 0.569 & 0.343 & 0.392 & 0.54 & 0.533 & 0.411 & \cellcolor{bronze}0.620 \\
DeltaCon & Euclidean & Spearman & \cellcolor{silver}0.738 & \cellcolor{bronze}0.580 & 0.566 & 0.325 & 0.379 & 0.539 & 0.518 & 0.417 & \cellcolor{gold}0.795 \\
DeltaCon & Euclidean & Xcorr & \cellcolor{gold}0.809 & \cellcolor{bronze}0.673 & 0.614 & 0.384 & 0.468 & 0.54 & 0.558 & 0.541 & \cellcolor{silver}0.807 \\
DeltaCon & Euclidean & DTW & 0.4 & 0.767 & \cellcolor{bronze}0.847 & 0.8 & 0.807 & 0.813 & \cellcolor{gold}0.940 & \cellcolor{silver}0.939 & 0.241 \\
Spectral & Cosine & Sym & \cellcolor{silver}0.941 & \cellcolor{bronze}0.415 & 0.172 & 0.35 & 0.344 & 0.333 & 0.019 & -0.004 & \cellcolor{gold}0.995 \\
Spectral & Cosine & Gauss & \cellcolor{silver}0.921 & 0.002 & \cellcolor{bronze}0.039 & 0.0 & 0.0 & 0.0 & 0.0 & 0.0 & \cellcolor{gold}0.992 \\
Spectral & Cosine & KL & \cellcolor{silver}0.944 & 0.128 & \cellcolor{bronze}0.177 & 0.059 & 0.043 & 0.046 & 0.0 & 0.0 & \cellcolor{gold}0.996 \\
Spectral & Cosine & Pearson & \cellcolor{gold}0.656 & 0.495 & 0.474 & 0.388 & 0.382 & \cellcolor{silver}0.583 & 0.528 & 0.403 & \cellcolor{bronze}0.572 \\
Spectral & Cosine & Spearman & \cellcolor{silver}0.653 & 0.532 & 0.507 & 0.401 & 0.377 & \cellcolor{bronze}0.580 & 0.522 & 0.398 & \cellcolor{gold}0.697 \\
Spectral & Cosine & Xcorr & \cellcolor{silver}0.667 & 0.521 & 0.542 & 0.441 & 0.464 & 0.583 & \cellcolor{bronze}0.585 & 0.507 & \cellcolor{gold}0.694 \\
Spectral & Cosine & DTW & \cellcolor{silver}0.941 & \cellcolor{bronze}0.415 & 0.172 & 0.35 & 0.344 & 0.333 & 0.019 & 0.0 & \cellcolor{gold}0.996 \\
Spectral & Euclidean & Sym & \cellcolor{silver}0.830 & \cellcolor{bronze}0.462 & 0.38 & 0.433 & 0.43 & 0.425 & 0.302 & 0.293 & \cellcolor{gold}0.989 \\
Spectral & Euclidean & Gauss & \cellcolor{silver}0.525 & 0.002 & \cellcolor{bronze}0.012 & 0.001 & 0.001 & 0.001 & 0.0 & 0.0 & \cellcolor{gold}0.992 \\
Spectral & Euclidean & KL & \cellcolor{silver}0.819 & \cellcolor{bronze}0.233 & 0.173 & 0.168 & 0.162 & 0.149 & 0.006 & 0.002 & \cellcolor{gold}0.991 \\
Spectral & Euclidean & Pearson & \cellcolor{gold}0.661 & 0.496 & 0.473 & 0.386 & 0.382 & \cellcolor{silver}0.583 & 0.529 & 0.407 & \cellcolor{bronze}0.582 \\
Spectral & Euclidean & Spearman & \cellcolor{silver}0.653 & 0.533 & 0.508 & 0.4 & 0.377 & \cellcolor{bronze}0.579 & 0.522 & 0.398 & \cellcolor{gold}0.710 \\
Spectral & Euclidean & Xcorr & \cellcolor{silver}0.670 & 0.523 & 0.537 & 0.443 & 0.469 & 0.583 & \cellcolor{bronze}0.586 & 0.51 & \cellcolor{gold}0.713 \\
Spectral & Euclidean & DTW & \cellcolor{silver}0.830 & \cellcolor{bronze}0.462 & 0.38 & 0.433 & 0.43 & 0.425 & 0.302 & 0.293 & \cellcolor{gold}0.989 \\
Euclidean & Cosine & Sym & 0.527 & \cellcolor{gold}0.947 & 0.603 & \cellcolor{silver}0.879 & \cellcolor{bronze}0.874 & 0.858 & 0.551 & 0.528 & 0.472 \\
Euclidean & Cosine & Gauss & 0.009 & \cellcolor{gold}0.931 & 0.328 & \cellcolor{silver}0.706 & \cellcolor{bronze}0.688 & 0.635 & 0.061 & 0.035 & 0.008 \\
Euclidean & Cosine & KL & 0.153 & \cellcolor{gold}0.993 & 0.532 & \cellcolor{bronze}0.948 & \cellcolor{silver}0.950 & 0.939 & 0.244 & 0.173 & 0.009 \\
Euclidean & Cosine & Pearson & \cellcolor{silver}0.871 & \cellcolor{gold}0.996 & \cellcolor{bronze}0.596 & 0.507 & 0.528 & 0.527 & 0.546 & 0.518 & 0.394 \\
Euclidean & Cosine & Spearman & \cellcolor{silver}0.857 & \cellcolor{gold}0.993 & \cellcolor{bronze}0.608 & 0.52 & 0.522 & 0.57 & 0.532 & 0.523 & 0.601 \\
Euclidean & Cosine & Xcorr & \cellcolor{silver}0.922 & \cellcolor{gold}0.998 & \cellcolor{bronze}0.606 & 0.537 & 0.53 & 0.547 & 0.546 & 0.518 & 0.586 \\
Euclidean & Cosine & DTW & 0.527 & \cellcolor{gold}0.969 & 0.604 & \cellcolor{bronze}0.894 & \cellcolor{silver}0.896 & 0.868 & 0.551 & 0.528 & 0.472 \\
Euclidean & Euclidean & Sym & 0.639 & \cellcolor{gold}0.986 & 0.843 & \cellcolor{silver}0.950 & \cellcolor{bronze}0.949 & 0.945 & 0.833 & 0.825 & 0.479 \\
Euclidean & Euclidean & Gauss & 0.062 & \cellcolor{gold}0.994 & 0.587 & \cellcolor{bronze}0.914 & \cellcolor{silver}0.915 & 0.904 & 0.544 & 0.52 & 0.007 \\
Euclidean & Euclidean & KL & 0.66 & \cellcolor{gold}0.999 & 0.902 & \cellcolor{bronze}0.991 & \cellcolor{silver}0.991 & 0.99 & 0.93 & 0.923 & 0.009 \\
Euclidean & Euclidean & Pearson & \cellcolor{silver}0.872 & \cellcolor{gold}0.996 & \cellcolor{bronze}0.585 & 0.509 & 0.53 & 0.527 & 0.541 & 0.52 & 0.409 \\
Euclidean & Euclidean & Spearman & \cellcolor{silver}0.857 & \cellcolor{gold}0.992 & 0.609 & 0.519 & 0.521 & 0.571 & 0.532 & 0.523 & \cellcolor{bronze}0.656 \\
Euclidean & Euclidean & Xcorr & \cellcolor{silver}0.923 & \cellcolor{gold}0.997 & 0.594 & 0.54 & 0.531 & 0.546 & 0.541 & 0.52 & \cellcolor{bronze}0.676 \\
Euclidean & Euclidean & DTW & 0.639 & \cellcolor{gold}0.995 & 0.848 & \cellcolor{bronze}0.959 & \cellcolor{silver}0.962 & 0.953 & 0.833 & 0.825 & 0.479 \\
\bottomrule
\end{tabular}

%% file: integrity_3move100_noalign.tex
\begin{tabular}{lllrrrrrrrrr}
\toprule
\multicolumn{3}{c}{}%
& \multicolumn{2}{c}{\textbf{Non-learning-based}}%
& \multicolumn{3}{c}{\textbf{Static}}%
& \multicolumn{4}{c}{\textbf{Dynamic}}\\
\cmidrule(lr){4-5}\cmidrule(lr){6-8}\cmidrule(lr){9-12}
$\Delta G$ & $\Delta R$ & kernel & IPP & UASE & GAT & AE & AE-Shuffled & DynAE & DynRNN & DynAERNN & STConv \\
\midrule
DeltaCon & Cosine & Sym & 0.256 & 0.686 & 0.646 & \cellcolor{gold}0.819 & \cellcolor{silver}0.814 & 0.795 & \cellcolor{bronze}0.809 & 0.806 & 0.217 \\
DeltaCon & Cosine & Gauss & 0.0 & 0.12 & 0.303 & \cellcolor{gold}0.493 & 0.473 & 0.412 & \cellcolor{silver}0.491 & \cellcolor{bronze}0.473 & 0.0 \\
DeltaCon & Cosine & KL & 0.002 & 0.788 & 0.452 & \cellcolor{gold}0.919 & \cellcolor{silver}0.915 & \cellcolor{bronze}0.898 & 0.468 & 0.47 & 0.006 \\
DeltaCon & Cosine & Pearson & \cellcolor{gold}0.832 & \cellcolor{silver}0.724 & \cellcolor{bronze}0.623 & 0.51 & 0.519 & 0.557 & 0.534 & 0.549 & 0.543 \\
DeltaCon & Cosine & Spearman & \cellcolor{gold}0.795 & 0.225 & \cellcolor{silver}0.646 & 0.477 & 0.511 & 0.554 & 0.515 & 0.545 & \cellcolor{bronze}0.602 \\
DeltaCon & Cosine & Xcorr & \cellcolor{gold}0.832 & \cellcolor{silver}0.724 & \cellcolor{bronze}0.660 & 0.557 & 0.609 & 0.597 & 0.553 & 0.549 & 0.645 \\
DeltaCon & Cosine & DTW & 0.256 & 0.686 & 0.67 & \cellcolor{bronze}0.819 & 0.814 & 0.795 & \cellcolor{silver}0.864 & \cellcolor{gold}0.876 & 0.217 \\
DeltaCon & Euclidean & Sym & 0.36 & 0.692 & \cellcolor{bronze}0.865 & 0.757 & 0.755 & 0.746 & \cellcolor{silver}0.901 & \cellcolor{gold}0.905 & 0.227 \\
DeltaCon & Euclidean & Gauss & 0.0 & 0.127 & \cellcolor{bronze}0.661 & 0.28 & 0.272 & 0.249 & \cellcolor{silver}0.782 & \cellcolor{gold}0.802 & 0.0 \\
DeltaCon & Euclidean & KL & 0.02 & 0.795 & \cellcolor{bronze}0.934 & 0.866 & 0.864 & 0.855 & \cellcolor{silver}0.969 & \cellcolor{gold}0.972 & 0.003 \\
DeltaCon & Euclidean & Pearson & \cellcolor{gold}0.846 & \cellcolor{silver}0.704 & \cellcolor{bronze}0.624 & 0.513 & 0.519 & 0.554 & 0.536 & 0.55 & 0.579 \\
DeltaCon & Euclidean & Spearman & \cellcolor{gold}0.797 & 0.223 & \cellcolor{bronze}0.646 & 0.478 & 0.508 & 0.548 & 0.515 & 0.547 & \cellcolor{silver}0.756 \\
DeltaCon & Euclidean & Xcorr & \cellcolor{gold}0.846 & \cellcolor{silver}0.704 & 0.659 & 0.556 & 0.606 & 0.596 & 0.554 & 0.55 & \cellcolor{bronze}0.666 \\
DeltaCon & Euclidean & DTW & 0.36 & 0.692 & \cellcolor{bronze}0.875 & 0.757 & 0.755 & 0.746 & \cellcolor{silver}0.903 & \cellcolor{gold}0.908 & 0.227 \\
Spectral & Cosine & Sym & \cellcolor{silver}0.950 & \cellcolor{bronze}0.520 & -0.004 & 0.386 & 0.392 & 0.411 & 0.024 & 0.014 & \cellcolor{gold}0.986 \\
Spectral & Cosine & Gauss & \cellcolor{silver}0.942 & \cellcolor{bronze}0.006 & 0.004 & 0.0 & 0.0 & 0.001 & 0.0 & 0.0 & \cellcolor{gold}0.987 \\
Spectral & Cosine & KL & \cellcolor{silver}0.954 & \cellcolor{bronze}0.352 & 0.067 & 0.074 & 0.084 & 0.128 & 0.001 & 0.0 & \cellcolor{gold}0.988 \\
Spectral & Cosine & Pearson & \cellcolor{gold}0.999 & \cellcolor{silver}0.967 & 0.521 & 0.542 & 0.511 & 0.53 & 0.531 & \cellcolor{bronze}0.544 & 0.511 \\
Spectral & Cosine & Spearman & \cellcolor{gold}0.636 & 0.352 & \cellcolor{bronze}0.547 & 0.423 & 0.506 & 0.473 & 0.434 & 0.418 & \cellcolor{silver}0.559 \\
Spectral & Cosine & Xcorr & \cellcolor{gold}0.999 & \cellcolor{silver}0.967 & 0.584 & 0.575 & \cellcolor{bronze}0.649 & 0.587 & 0.598 & 0.544 & 0.579 \\
Spectral & Cosine & DTW & \cellcolor{silver}0.950 & \cellcolor{bronze}0.520 & 0.0 & 0.386 & 0.392 & 0.411 & 0.024 & 0.014 & \cellcolor{gold}0.986 \\
Spectral & Euclidean & Sym & \cellcolor{silver}0.846 & \cellcolor{bronze}0.514 & 0.307 & 0.449 & 0.451 & 0.46 & 0.304 & 0.301 & \cellcolor{gold}0.977 \\
Spectral & Euclidean & Gauss & \cellcolor{silver}0.590 & \cellcolor{bronze}0.005 & 0.002 & 0.001 & 0.001 & 0.002 & 0.0 & 0.0 & \cellcolor{gold}0.982 \\
Spectral & Euclidean & KL & \cellcolor{silver}0.839 & \cellcolor{bronze}0.343 & 0.083 & 0.206 & 0.211 & 0.23 & 0.008 & 0.003 & \cellcolor{gold}0.980 \\
Spectral & Euclidean & Pearson & \cellcolor{gold}0.997 & \cellcolor{silver}0.955 & 0.525 & 0.543 & 0.513 & 0.531 & 0.532 & 0.544 & \cellcolor{bronze}0.547 \\
Spectral & Euclidean & Spearman & \cellcolor{silver}0.640 & 0.353 & \cellcolor{bronze}0.548 & 0.422 & 0.508 & 0.473 & 0.434 & 0.42 & \cellcolor{gold}0.641 \\
Spectral & Euclidean & Xcorr & \cellcolor{gold}0.997 & \cellcolor{silver}0.955 & 0.576 & 0.574 & \cellcolor{bronze}0.644 & 0.586 & 0.593 & 0.544 & 0.547 \\
Spectral & Euclidean & DTW & \cellcolor{silver}0.846 & \cellcolor{bronze}0.514 & 0.307 & 0.449 & 0.451 & 0.46 & 0.304 & 0.301 & \cellcolor{gold}0.977 \\
Euclidean & Cosine & Sym & 0.693 & \cellcolor{gold}0.877 & 0.351 & 0.744 & \cellcolor{bronze}0.749 & \cellcolor{silver}0.768 & 0.381 & 0.371 & 0.654 \\
Euclidean & Cosine & Gauss & 0.125 & \cellcolor{gold}0.713 & 0.137 & 0.248 & \cellcolor{bronze}0.262 & \cellcolor{silver}0.326 & 0.007 & 0.001 & 0.078 \\
Euclidean & Cosine & KL & 0.545 & \cellcolor{gold}0.969 & 0.318 & 0.861 & \cellcolor{bronze}0.867 & \cellcolor{silver}0.884 & 0.125 & 0.065 & 0.041 \\
Euclidean & Cosine & Pearson & \cellcolor{silver}0.976 & \cellcolor{gold}0.994 & 0.498 & \cellcolor{bronze}0.552 & 0.516 & 0.524 & 0.536 & 0.536 & 0.489 \\
Euclidean & Cosine & Spearman & 0.253 & \cellcolor{gold}0.966 & 0.444 & \cellcolor{silver}0.564 & \cellcolor{bronze}0.550 & 0.53 & 0.518 & 0.452 & 0.391 \\
Euclidean & Cosine & Xcorr & \cellcolor{silver}0.976 & \cellcolor{gold}0.994 & 0.552 & 0.582 & \cellcolor{bronze}0.649 & 0.576 & 0.598 & 0.536 & 0.531 \\
Euclidean & Cosine & DTW & 0.693 & \cellcolor{gold}0.877 & 0.471 & 0.771 & \cellcolor{bronze}0.782 & \cellcolor{silver}0.811 & 0.386 & 0.371 & 0.655 \\
Euclidean & Euclidean & Sym & 0.797 & \cellcolor{gold}0.871 & 0.665 & 0.805 & \cellcolor{bronze}0.807 & \cellcolor{silver}0.816 & 0.661 & 0.658 & 0.664 \\
Euclidean & Euclidean & Gauss & 0.402 & \cellcolor{gold}0.690 & 0.207 & 0.432 & \cellcolor{bronze}0.441 & \cellcolor{silver}0.475 & 0.098 & 0.086 & 0.089 \\
Euclidean & Euclidean & KL & 0.877 & \cellcolor{gold}0.966 & 0.675 & 0.922 & \cellcolor{bronze}0.924 & \cellcolor{silver}0.930 & 0.748 & 0.746 & 0.263 \\
Euclidean & Euclidean & Pearson & \cellcolor{silver}0.966 & \cellcolor{gold}0.990 & 0.503 & \cellcolor{bronze}0.552 & 0.517 & 0.524 & 0.535 & 0.536 & 0.516 \\
Euclidean & Euclidean & Spearman & 0.252 & \cellcolor{gold}0.967 & 0.445 & \cellcolor{silver}0.566 & \cellcolor{bronze}0.549 & 0.533 & 0.518 & 0.452 & 0.183 \\
Euclidean & Euclidean & Xcorr & \cellcolor{silver}0.966 & \cellcolor{gold}0.990 & 0.546 & 0.58 & \cellcolor{bronze}0.644 & 0.574 & 0.592 & 0.536 & 0.516 \\
Euclidean & Euclidean & DTW & 0.797 & \cellcolor{gold}0.871 & 0.77 & 0.806 & \cellcolor{bronze}0.809 & \cellcolor{silver}0.818 & 0.685 & 0.658 & 0.664 \\
\bottomrule
\end{tabular}

%% file: integrity_periodic100_noalign.tex
\begin{tabular}{lllrrrrrrrrr}
\toprule
\multicolumn{3}{c}{}%
& \multicolumn{2}{c}{\textbf{Non-learning-based}}%
& \multicolumn{3}{c}{\textbf{Static}}%
& \multicolumn{4}{c}{\textbf{Dynamic}}\\
\cmidrule(lr){4-5}\cmidrule(lr){6-8}\cmidrule(lr){9-12}
$\Delta G$ & $\Delta R$ & kernel & IPP & UASE & GAT & AE & AE-Shuffled & DynAE & DynRNN & DynAERNN & STConv \\
\midrule
DeltaCon & Cosine & Sym & 0.316 & \cellcolor{gold}0.834 & 0.642 & 0.802 & 0.792 & 0.726 & \cellcolor{bronze}0.805 & \cellcolor{silver}0.807 & 0.205 \\
DeltaCon & Cosine & Gauss & 0.0 & \cellcolor{gold}0.558 & 0.238 & 0.436 & 0.398 & 0.235 & \cellcolor{bronze}0.487 & \cellcolor{silver}0.497 & 0.0 \\
DeltaCon & Cosine & KL & 0.021 & \cellcolor{gold}0.910 & 0.435 & \cellcolor{silver}0.902 & \cellcolor{bronze}0.894 & 0.824 & 0.438 & 0.448 & 0.003 \\
DeltaCon & Cosine & Pearson & \cellcolor{silver}0.892 & \cellcolor{gold}0.905 & 0.456 & 0.481 & 0.448 & 0.427 & 0.438 & 0.54 & \cellcolor{bronze}0.584 \\
DeltaCon & Cosine & Spearman & \cellcolor{gold}0.913 & \cellcolor{silver}0.894 & 0.462 & 0.47 & 0.454 & 0.409 & 0.446 & 0.501 & \cellcolor{bronze}0.860 \\
DeltaCon & Cosine & Xcorr & \cellcolor{silver}0.892 & \cellcolor{gold}0.905 & 0.57 & 0.611 & 0.531 & 0.557 & 0.571 & 0.55 & \cellcolor{bronze}0.797 \\
DeltaCon & Cosine & DTW & 0.316 & \cellcolor{gold}0.859 & 0.659 & 0.815 & 0.792 & 0.726 & \cellcolor{bronze}0.833 & \cellcolor{silver}0.841 & 0.205 \\
DeltaCon & Euclidean & Sym & 0.432 & 0.752 & \cellcolor{bronze}0.802 & 0.741 & 0.737 & 0.706 & \cellcolor{silver}0.893 & \cellcolor{gold}0.894 & 0.232 \\
DeltaCon & Euclidean & Gauss & 0.001 & 0.273 & \cellcolor{bronze}0.519 & 0.24 & 0.227 & 0.165 & \cellcolor{silver}0.752 & \cellcolor{gold}0.762 & 0.0 \\
DeltaCon & Euclidean & KL & 0.264 & 0.859 & \cellcolor{bronze}0.862 & 0.848 & 0.843 & 0.807 & \cellcolor{silver}0.963 & \cellcolor{gold}0.965 & 0.002 \\
DeltaCon & Euclidean & Pearson & \cellcolor{silver}0.902 & \cellcolor{gold}0.907 & 0.452 & 0.479 & 0.448 & 0.43 & 0.439 & 0.54 & \cellcolor{bronze}0.657 \\
DeltaCon & Euclidean & Spearman & \cellcolor{gold}0.913 & \cellcolor{silver}0.895 & 0.462 & 0.469 & 0.452 & 0.409 & 0.446 & 0.501 & \cellcolor{bronze}0.880 \\
DeltaCon & Euclidean & Xcorr & \cellcolor{silver}0.902 & \cellcolor{gold}0.907 & 0.55 & 0.612 & 0.533 & 0.555 & 0.571 & 0.554 & \cellcolor{bronze}0.802 \\
DeltaCon & Euclidean & DTW & 0.432 & 0.752 & \cellcolor{bronze}0.828 & 0.741 & 0.737 & 0.706 & \cellcolor{silver}0.918 & \cellcolor{gold}0.926 & 0.232 \\
Spectral & Cosine & Sym & \cellcolor{silver}0.887 & 0.368 & 0.17 & 0.401 & 0.411 & \cellcolor{bronze}0.476 & 0.017 & 0.012 & \cellcolor{gold}0.988 \\
Spectral & Cosine & Gauss & \cellcolor{silver}0.733 & 0.003 & \cellcolor{bronze}0.040 & 0.001 & 0.001 & 0.005 & 0.0 & 0.0 & \cellcolor{gold}0.990 \\
Spectral & Cosine & KL & \cellcolor{silver}0.903 & 0.154 & 0.197 & 0.128 & 0.151 & \cellcolor{bronze}0.289 & 0.0 & 0.0 & \cellcolor{gold}0.993 \\
Spectral & Cosine & Pearson & \cellcolor{silver}0.646 & \cellcolor{gold}0.667 & 0.531 & \cellcolor{bronze}0.598 & 0.463 & 0.414 & 0.547 & 0.437 & 0.57 \\
Spectral & Cosine & Spearman & \cellcolor{silver}0.609 & 0.567 & 0.515 & \cellcolor{bronze}0.595 & 0.453 & 0.424 & 0.53 & 0.419 & \cellcolor{gold}0.648 \\
Spectral & Cosine & Xcorr & \cellcolor{bronze}0.646 & \cellcolor{silver}0.667 & 0.531 & 0.598 & 0.493 & 0.514 & 0.547 & 0.571 & \cellcolor{gold}0.667 \\
Spectral & Cosine & DTW & \cellcolor{silver}0.887 & 0.368 & 0.17 & 0.401 & 0.411 & \cellcolor{bronze}0.476 & 0.017 & 0.012 & \cellcolor{gold}0.991 \\
Spectral & Euclidean & Sym & \cellcolor{silver}0.770 & 0.45 & 0.386 & 0.461 & 0.466 & \cellcolor{bronze}0.497 & 0.309 & 0.307 & \cellcolor{gold}0.970 \\
Spectral & Euclidean & Gauss & \cellcolor{silver}0.339 & 0.003 & \cellcolor{bronze}0.013 & 0.002 & 0.002 & 0.005 & 0.0 & 0.0 & \cellcolor{gold}0.974 \\
Spectral & Euclidean & KL & \cellcolor{silver}0.770 & 0.232 & 0.193 & 0.264 & 0.274 & \cellcolor{bronze}0.337 & 0.013 & 0.009 & \cellcolor{gold}0.980 \\
Spectral & Euclidean & Pearson & \cellcolor{silver}0.661 & \cellcolor{gold}0.673 & 0.525 & \cellcolor{bronze}0.597 & 0.462 & 0.415 & 0.545 & 0.44 & 0.594 \\
Spectral & Euclidean & Spearman & \cellcolor{gold}0.610 & 0.567 & 0.516 & \cellcolor{bronze}0.596 & 0.451 & 0.424 & 0.53 & 0.419 & \cellcolor{silver}0.609 \\
Spectral & Euclidean & Xcorr & \cellcolor{silver}0.661 & \cellcolor{gold}0.673 & 0.525 & 0.597 & 0.495 & 0.516 & 0.545 & 0.571 & \cellcolor{bronze}0.624 \\
Spectral & Euclidean & DTW & \cellcolor{silver}0.770 & 0.45 & 0.386 & 0.461 & 0.466 & \cellcolor{bronze}0.497 & 0.309 & 0.307 & \cellcolor{gold}0.973 \\
Euclidean & Cosine & Sym & 0.546 & \cellcolor{gold}0.935 & 0.592 & \cellcolor{bronze}0.907 & \cellcolor{silver}0.913 & 0.88 & 0.584 & 0.579 & 0.436 \\
Euclidean & Cosine & Gauss & 0.014 & \cellcolor{gold}0.882 & 0.262 & \cellcolor{bronze}0.791 & \cellcolor{silver}0.809 & 0.706 & 0.12 & 0.082 & 0.004 \\
Euclidean & Cosine & KL & 0.387 & \cellcolor{gold}0.984 & 0.481 & \cellcolor{bronze}0.975 & \cellcolor{silver}0.977 & 0.962 & 0.27 & 0.26 & 0.006 \\
Euclidean & Cosine & Pearson & \cellcolor{silver}0.994 & \cellcolor{gold}1.000 & 0.501 & 0.52 & 0.539 & 0.362 & 0.445 & 0.577 & \cellcolor{bronze}0.629 \\
Euclidean & Cosine & Spearman & \cellcolor{gold}0.993 & \cellcolor{silver}0.991 & 0.508 & 0.492 & 0.516 & 0.373 & 0.449 & 0.569 & \cellcolor{bronze}0.882 \\
Euclidean & Cosine & Xcorr & \cellcolor{silver}0.994 & \cellcolor{gold}1.000 & 0.512 & 0.565 & 0.539 & 0.586 & 0.578 & 0.577 & \cellcolor{bronze}0.793 \\
Euclidean & Cosine & DTW & 0.546 & \cellcolor{gold}0.946 & 0.648 & \cellcolor{bronze}0.942 & \cellcolor{silver}0.946 & 0.938 & 0.61 & 0.579 & 0.436 \\
Euclidean & Euclidean & Sym & 0.663 & \cellcolor{gold}0.966 & 0.833 & \cellcolor{silver}0.909 & \cellcolor{bronze}0.909 & 0.89 & 0.863 & 0.872 & 0.463 \\
Euclidean & Euclidean & Gauss & 0.09 & \cellcolor{gold}0.967 & 0.56 & \cellcolor{silver}0.806 & \cellcolor{bronze}0.805 & 0.736 & 0.64 & 0.663 & 0.007 \\
Euclidean & Euclidean & KL & 0.735 & \cellcolor{gold}0.997 & 0.905 & \cellcolor{silver}0.978 & \cellcolor{bronze}0.978 & 0.967 & 0.937 & 0.942 & 0.036 \\
Euclidean & Euclidean & Pearson & \cellcolor{gold}0.999 & \cellcolor{silver}0.999 & 0.492 & 0.517 & 0.538 & 0.365 & 0.442 & 0.58 & \cellcolor{bronze}0.727 \\
Euclidean & Euclidean & Spearman & \cellcolor{gold}0.992 & \cellcolor{silver}0.991 & 0.508 & 0.49 & 0.515 & 0.374 & 0.449 & 0.569 & \cellcolor{bronze}0.956 \\
Euclidean & Euclidean & Xcorr & \cellcolor{gold}0.999 & \cellcolor{silver}0.999 & 0.529 & 0.566 & 0.538 & 0.584 & 0.577 & 0.58 & \cellcolor{bronze}0.861 \\
Euclidean & Euclidean & DTW & 0.663 & \cellcolor{gold}0.966 & 0.89 & \cellcolor{bronze}0.919 & 0.916 & \cellcolor{silver}0.921 & 0.904 & 0.893 & 0.463 \\
\bottomrule
\end{tabular}

%% file: wins.tex
\begin{table}[htbp]
  \centering
  \caption{Integrity-gap and win statistics for the chosen index (Euclidean, Euclidean, Pearson) in both not aligned and aligned versions. The first column shows if representation alignment is performed or not.}
  \label{tab:wins}
  \begin{tabular}{lccc}
    \hline
    Alignment & $\Delta \bar{I}_{\text{UASE}}$
    & $\Delta \bar{I}_{\text{IPP}}$
    & $\text{Win}_{\text{UASE}|\text{IPP}}$ \\
    \hline
    no
    & 0.502 & 0.466 & 3 \\
   yes
    & 0.468 & 0.431 & 3 \\
    \hline
  \end{tabular}
\end{table}

%% file: comparison0.tex
\begin{table}[ht]
  \centering
  \setlength\tabcolsep{5pt}
  \renewcommand{\arraystretch}{1.1}
  \caption[Integrity scores for each model on the three synthetic scenarios]{Integrity scores for each model on the three synthetic scenarios, computed with the validated Euclidean–Euclidean–Pearson integrity index. Models are grouped by learning paradigm; the Alignment column indicates whether representation alignment was applied (Yes) or not (No) before the metric was computed.}
  \begin{adjustbox}{max width=\linewidth}
  \begin{tabular}{|c|c|c|c|c|c|}
    \hline
    \multirow{2}{*}{\textbf{Alignment}} &
    \multirow{2}{*}{\textbf{Category}} &
    \multirow{2}{*}{\textbf{Model}} &
    \multicolumn{3}{c|}{\textbf{Dataset}} \\ \cline{4-6}
    & & & \textit{Merge} & \textit{Move} & \textit{Periodic} \\ \hline
    \multirow{9}{*}{No} & \multirow{2}{*}{Non learning-based}
      & IPP  & \cellcolor{silver}0.872 & \cellcolor{silver}0.966 & \cellcolor{gold}0.999 \\ \cline{3-6}
    & & UASE & \cellcolor{gold}0.996   & \cellcolor{gold}0.990   & \cellcolor{silver}0.999 \\ \cline{2-6}
    & \multirow{3}{*}{Static}
      & GAT          & \cellcolor{bronze}0.585 & 0.503 & 0.492 \\ \cline{3-6}
    & & AE           & 0.509 & \cellcolor{bronze}0.552 & 0.517 \\ \cline{3-6}
    & & AE\_Shuffled & 0.530 & 0.517 & 0.538 \\ \cline{2-6}
    & \multirow{4}{*}{Dynamic}
      & DynAE    & 0.527 & 0.524 & 0.365 \\ \cline{3-6}
    & & DynRNN   & 0.541 & 0.535 & 0.442 \\ \cline{3-6}
    & & DynAERNN & 0.520 & 0.536 & 0.580 \\ \cline{3-6}
    & & STConv   & 0.409 & 0.516 & \cellcolor{bronze}0.727 \\ \hline
    \multirow{9}{*}{Yes} & \multirow{2}{*}{Non learning-based}
      & IPP  & \cellcolor{silver}0.866 & \cellcolor{gold}0.890 & \cellcolor{gold}0.999 \\ \cline{3-6}
    & & UASE & \cellcolor{gold}0.995 & \cellcolor{silver}0.874 & \cellcolor{silver}0.998 \\ \cline{2-6}
    & \multirow{3}{*}{Static}
      & GAT          & 0.45 & 0.467 & 0.542 \\ \cline{3-6}
    & & AE           & \cellcolor{bronze}0.699 & 0.442 & 0.961 \\ \cline{3-6}
    & & AE\_Shuffled & 0.581 & 0.512 & 0.425 \\ \cline{2-6}
    & \multirow{4}{*}{Dynamic}
      & DynAE    & 0.552 & 0.404 & \cellcolor{bronze}0.965 \\ \cline{3-6}
    & & DynRNN   & 0.522 & 0.459 & 0.556 \\ \cline{3-6}
    & & DynAERNN & 0.318 & \cellcolor{bronze}0.627 & 0.427 \\ \cline{3-6}
    & & STConv   & 0.411 & 0.52 & 0.738 \\ \hline
  \end{tabular}
  \end{adjustbox}
  \label{tab:comparison}
\end{table}

%% file: Chapters/conclusion.tex
\chapter{Discussions and Conclusion}

This thesis set out to answer a fundamental question: \textit{How can we tell whether a dynamic graph–learning model is faithfully capturing the evolving structure of a network, irrespective of any downstream task?}  
To that end, we introduced the notion of \emph{representation integrity}, formalised a modular design space of integrity indexes, and screened forty-two candidate metrics against a principled set of properties and synthetic tests.  
The outcome is a single, validated index-Euclidean graph change, Euclidean representation change, Pearson alignment-that we then used to benchmark eight representative embedding models on three synthetic scenarios and a real parliamentary-voting dataset.  
This chapter interprets those results, situates them in the broader literature, and reflects on the practical and theoretical implications of integrity-oriented evaluation.

Prior work on dynamic embeddings has largely relied on task performance-link prediction, node classification, or anomaly detection-as a proxy for quality.  
While convenient, such proxies conflate representational faithfulness with task-specific inductive biases.  
Our first contribution was therefore conceptual: we framed integrity as a task-agnostic, bidirectional coupling between graph dynamics and embedding dynamics.  
Building on that foundation, we argued that any integrity index must (i) compare graph change to representation change in the \emph{same} time window, and (ii) account for arbitrary rigid motions through an alignment kernel.

The methodological pipeline that followed-property selection, synthetic scenario design, exhaustive metric screening-was motivated by a need for \emph{construct validity}.  
Only an index that passes all properties and reproduces known ground-truth relationships in controlled settings can meaningfully adjudicate between models on real data.

The exhaustive search revealed that most intuitive combinations of distance measures fail at least one property or scenario.  
Our chosen index, the Euclidean–Euclidean–Pearson triplet, succeeds because (a) Euclidean distance linearises small perturbations in both the graph and embedding spaces, and (b) Pearson correlation removes scale dependence introduced by alignment.  

\subsection*{Model Comparison}

Across all scenarios, non-learning based models consistently outperform others, with or without procrustes alignment performed on the representation. Without the alignment, we see that STConv is the only model that achieves a considerably high result. We see that dynamic models unfortunately don't outperform static models generally. However, the alignment procedure helps Auto Encoder based models like AE, DynAE and DynAERNN to achieve much better integrity scores.

\subsection*{Integrity as a Proxy for Task Performance}

A strong average Spearman correlation ($\rho = 0.596$) between integrity and one-step link-prediction AUC on the synthetic datasets shows that integrity has a positive correlation with downstream effectiveness.  

\section{Implications}\label{sec:discussion-implications}
 
Temporal convolution (or any operation that couples multiple snapshots in the feature space) emerges as a first-class design principle.  
Architectures that treat time as a mere sequence of static graphs risk overlooking long-range correlations that integrity penalizes and downstream tasks implicitly reward.
 
The validated index provides a lightweight, task-independent diagnostic tool.  
Practitioners can now evaluate candidate models on \emph{their own} data before committing to the heavy engineering effort of task-specific fine-tuning. In practice, integrity scores can guide concrete decisions: models with low integrity may be ruled out early, while models with high integrity can be prioritized for further optimization.  

This allows practitioners to allocate computational budgets more effectively, choose architectures that are more likely to remain robust under distribution shifts, and build greater confidence in model stability before deployment.  
 
Our property-based framework opens two research directions: (i) deriving tighter theoretical bounds on integrity for classes of graph processes; and (ii) designing learning objectives that maximize integrity explicitly, analogous to how contrastive losses maximize mutual information.

\section{Limitations}\label{sec:discussion-limitations}

Three caveats temper our conclusions.  
First, the synthetic scenarios, while diverse, cannot span the full variability of real-world dynamics; integrity behavior under adversarial or highly non-stationary conditions remains an open question.  
Second, the benchmark set covers only eight models; emerging techniques such as continuous-time diffusion-based embeddings were beyond our computational budget.  
Third, we validated a single index, but alternative graph distances or alignment kernels may prove suitable if paired with different properties-a line worth exploring.

\section{Future Work}\label{sec:discussion-future}

Two strands of future research appear particularly promising:

\begin{enumerate}
  \item \textbf{Integrity-aware training.} Incorporating the validated index (or a differentiable surrogate) into the loss function could steer models toward embeddings that are both predictive and faithful.
  \item \textbf{Domain-specific scenarios.} Crafting stress tests that mimic domain-typical phenomena-e.g.\ churn in social networks, bursts in communication graphs-would sharpen the diagnostic power of integrity beyond generic synthetic benchmarks.
\end{enumerate}

\section{Conclusion}\label{sec:discussion-conclusion}

By disentangling representational faithfulness from task performance, this thesis provides a principled yardstick-\emph{representation integrity}-for the rapidly evolving field of dynamic graph learning.  
Our empirical study demonstrates that integrity both discriminates meaningfully among models and correlates with predictive success, making it an actionable criterion for model selection.  
The broader lesson is that good representations \emph{move} in ways that mirror their data: respecting that motion is not a luxury but a necessity for robust, generalizable learning.

%% file: Chapters/appendix.tex
\chapter{}
\label{app:appendix}

\section{Index validation - Integrity gap and win statistics for all indexes before and after Procrustes Alignment}

Tables~\ref{tab:gap-wins1} and ~\ref{tab:gap-wins2} present the validation results for all indexes where representations are either used without modification or aligned before the integrity score computation, respectively.

\input{integrity_gap_wins_nonaligned.tex}

\input{integrity_gap_wins.tex}

\section{Integrity Scores after Procrustes Alignment}

After performing Procrustes alignment on the representations, we compute integrity scores and report them in Tables \ref{tab:integrity-merge-aligned}, \ref{tab:integrity-move-aligned}, and \ref{tab:integrity-periodic-aligned} for merge, move, and periodic settings, respectively.

\begin{table}[t] \centering \caption{Integrity scores in merge setting, when representations are aligned between timestamps.} \begin{adjustbox}{max width=\linewidth, max height=0.9\textheight, center} \input{integrity_2merge100_aligned.tex} \end{adjustbox} \label{tab:integrity-merge-aligned} \end{table}

\begin{table}[t] \centering \caption{Integrity scores in move setting.} \begin{adjustbox}{max width=\linewidth, max height=0.9\textheight, center} \input{integrity_3move100_aligned.tex} \end{adjustbox} \label{tab:integrity-move-aligned} \end{table}

\begin{table}[t] \centering \caption{Integrity scores in periodic setting.} \begin{adjustbox}{max width=\linewidth, max height=0.9\textheight, center} \input{integrity_periodic100_aligned.tex} \end{adjustbox} \label{tab:integrity-periodic-aligned} \end{table}

In Table~\ref{tab:aligned_euc_euc_pearson}, we present the integrity scores of various models based on the validated integrity index - Euclidean, Euclidean and Pearson composition. In the merge setting, UASE achieves the highest integrity score of 0.995, slightly ahead of IPP. All other learned models fall significantly behind, with AE scoring 0.699, the best among the others. In the move setting, IPP (0.890) surpasses UASE, while the only learned model that narrows the gap is DynAERNN (0.627), indicating that its recurrent attention mechanism effectively tracks the shifting community. In the periodic setting, IPP narrowly outperforms UASE (0.999 versus 0.998), but DynAE stands out among the neural approaches with a score of 0.965. 

When comparing integrity scores before alignment in Table~\ref{tab:comparison}, IPP and UASE consistently rank as the top two models. However, the scores for AE and DynAE show a significant improvement, whereas the scores for other models do not show as drastic a boost.

\begin{table}[ht]
  \centering
  \setlength\tabcolsep{5pt}
  \renewcommand{\arraystretch}{1.1}
  \caption{Integrity scores for each model after representation
           alignment.}
  \label{tab:aligned_euc_euc_pearson}
    \begin{adjustbox}{max width=\linewidth}
     \begin{tabular}{|c|c|c|c|c|}
    \hline
    \multirow{2}{*}{\textbf{Category}} & \multirow{2}{*}{\textbf{Model}}
      & \multicolumn{3}{c|}{\textbf{Dataset}} \\ \cline{3-5}
      & & \textit{Merge} & \textit{Move} & \textit{Periodic} \\ \hline
    \multirow{2}{*}{Non learning-based} 
      & IPP        & \cellcolor{silver}0.866 & \cellcolor{gold}0.890 & \cellcolor{gold}0.999 \\ \cline{2-5}
    & UASE       & \cellcolor{gold}0.995 & \cellcolor{silver}0.874 & \cellcolor{silver}0.998 \\ \hline
    \multirow{3}{*}{Static}
    & GAT       & 0.450 & 0.467 & 0.542 \\
    \cline{2-5}
    & AE         & \cellcolor{bronze}0.699 & 0.442 & 0.961 \\ \cline{2-5}
    & AE-Shuffled & 0.581 & 0.512 & 0.425 \\ \hline
    \multirow{3}{*}{Dynamic}
    & DynAE      & 0.552 & 0.404 & \cellcolor{bronze}0.965 \\  \cline{2-5}
    & DynRNN     & 0.522 & 0.459 & 0.556 \\  \cline{2-5}
    & DynAERNN   & 0.318 & \cellcolor{bronze}0.627 & 0.427 \\ \cline{2-5}
    & STConv     & 0.411 & 0.520 & 0.738 \\
    \hline
  \end{tabular}
  \end{adjustbox}
\end{table}

%% file: integrity_gap_wins_nonaligned.tex
\begin{longtable}{lllrrr}
\caption{Integrity‐gap and win statistics for every index in nonaligned version.}%
\label{tab:gap-wins1}\\
\toprule
$\Delta G$ & $\Delta R$ & kernel &
$\Delta \bar{I}_{\text{UASE}}$ &
$\Delta \bar{I}_{\text{IPP}}$ &
Win$_{\text{UASE|IPP}}$ \\
\midrule
\endfirsthead

\caption[]{Integrity‐gap and win statistics (continued).}\\
\toprule
$\Delta G$ & $\Delta R$ & kernel &
$\Delta \bar{I}_{\text{UASE}}$ &
$\Delta \bar{I}_{\text{IPP}}$ &
Win$_{\text{UASE|IPP}}$ \\
\midrule
\endhead

\midrule
\multicolumn{6}{r}{\emph{Continued on next page}}\\
\midrule
\endfoot

\bottomrule
\endlastfoot
Euclidean & Euclidean & DTW & 0.048 & -0.237 & 3 \\
Euclidean & Cosine & DTW & 0.086 & -0.289 & 3 \\
Spectral & Euclidean & Spearman & 0.003 & 0.147 & 3 \\
Spectral & Euclidean & Pearson & 0.158 & 0.264 & 3 \\
Euclidean & Cosine & Sym & 0.088 & -0.275 & 3 \\
Euclidean & Cosine & Gauss & 0.305 & -0.576 & 3 \\
Euclidean & Cosine & KL & 0.062 & -0.617 & 3 \\
Euclidean & Cosine & Pearson & 0.499 & 0.472 & 3 \\
Spectral & Cosine & Xcorr & 0.121 & 0.228 & 3 \\
Spectral & Cosine & Spearman & 0.003 & 0.147 & 3 \\
Spectral & Cosine & Pearson & 0.152 & 0.263 & 3 \\
Euclidean & Cosine & Spearman & 0.548 & 0.271 & 3 \\
Euclidean & Cosine & Xcorr & 0.435 & 0.424 & 3 \\
Euclidean & Euclidean & Sym & 0.051 & -0.231 & 3 \\
DeltaCon & Euclidean & Xcorr & 0.112 & 0.281 & 3 \\
DeltaCon & Euclidean & Spearman & 0.034 & 0.280 & 3 \\
DeltaCon & Euclidean & Pearson & 0.152 & 0.332 & 3 \\
Euclidean & Euclidean & Gauss & 0.132 & -0.693 & 3 \\
Euclidean & Euclidean & KL & 0.016 & -0.243 & 3 \\
Euclidean & Euclidean & Pearson & 0.502 & 0.466 & 3 \\
Euclidean & Euclidean & Spearman & 0.548 & 0.273 & 3 \\
DeltaCon & Cosine & Xcorr & 0.106 & 0.272 & 3 \\
DeltaCon & Cosine & Spearman & 0.033 & 0.282 & 3 \\
DeltaCon & Cosine & Pearson & 0.146 & 0.323 & 3 \\
Euclidean & Euclidean & Xcorr & 0.438 & 0.418 & 3 \\
Spectral & Euclidean & Xcorr & 0.129 & 0.229 & 3 \\
Spectral & Euclidean & DTW & -0.037 & 0.281 & 2 \\
DeltaCon & Cosine & Sym & 0.049 & -0.392 & 2 \\
Spectral & Euclidean & KL & -0.054 & 0.441 & 2 \\
Spectral & Euclidean & Gauss & -0.059 & 0.454 & 2 \\
DeltaCon & Cosine & Gauss & -0.007 & -0.360 & 2 \\
Spectral & Cosine & DTW & -0.049 & 0.392 & 2 \\
Spectral & Cosine & KL & -0.045 & 0.589 & 2 \\
Spectral & Cosine & Gauss & -0.170 & 0.666 & 2 \\
Spectral & Cosine & Sym & -0.048 & 0.393 & 2 \\
DeltaCon & Cosine & DTW & 0.046 & -0.407 & 2 \\
DeltaCon & Cosine & KL & 0.156 & -0.610 & 2 \\
Spectral & Euclidean & Sym & -0.037 & 0.281 & 2 \\
DeltaCon & Euclidean & DTW & 0.036 & -0.282 & 1 \\
DeltaCon & Euclidean & KL & 0.086 & -0.578 & 1 \\
DeltaCon & Euclidean & Gauss & -0.007 & -0.215 & 1 \\
DeltaCon & Euclidean & Sym & 0.037 & -0.281 & 1 \\
\bottomrule
\end{longtable}

%% file: integrity_gap_wins.tex
\begin{longtable}{lllrrr}
\caption{Integrity‐gap and win statistics for every index in aligned version.}%
\label{tab:gap-wins2}\\
\toprule
$\Delta G$ & $\Delta R$ & kernel &
$\Delta \bar{I}_{\text{UASE}}$ &
$\Delta \bar{I}_{\text{IPP}}$ &
Win$_{\text{UASE|IPP}}$ \\
\midrule
\endfirsthead

\caption[]{Integrity‐gap and win statistics (continued).}\\
\toprule
$\Delta G$ & $\Delta R$ & kernel &
$\Delta \bar{I}_{\text{UASE}}$ &
$\Delta \bar{I}_{\text{IPP}}$ &
Win$_{\text{UASE|IPP}}$ \\
\midrule
\endhead

\midrule
\multicolumn{6}{r}{\emph{Continued on next page}}\\
\midrule
\endfoot

\bottomrule
\endlastfoot
DeltaCon & Cosine & Sym & 0.214 & -0.099 & 3 \\
DeltaCon & Cosine & Gauss & 0.029 & 0.000 & 3 \\
Euclidean & Euclidean & Xcorr & 0.435 & 0.414 & 3 \\
Euclidean & Euclidean & Spearman & 0.548 & 0.271 & 3 \\
Euclidean & Euclidean & Pearson & 0.468 & 0.431 & 3 \\
Euclidean & Euclidean & KL & 0.117 & -0.189 & 3 \\
Euclidean & Euclidean & Gauss & 0.382 & -0.294 & 3 \\
Euclidean & Euclidean & Sym & 0.130 & -0.117 & 3 \\
Euclidean & Cosine & DTW & 0.223 & -0.100 & 3 \\
Euclidean & Cosine & Xcorr & 0.435 & 0.424 & 3 \\
Euclidean & Cosine & Spearman & 0.548 & 0.269 & 3 \\
Euclidean & Cosine & Pearson & 0.470 & 0.442 & 3 \\
Euclidean & Cosine & KL & 0.317 & -0.450 & 3 \\
Euclidean & Cosine & Gauss & 0.516 & -0.123 & 3 \\
Euclidean & Cosine & Sym & 0.213 & -0.100 & 3 \\
Euclidean & Euclidean & DTW & 0.141 & -0.117 & 3 \\
DeltaCon & Euclidean & DTW & 0.154 & -0.116 & 3 \\
DeltaCon & Euclidean & Gauss & 0.056 & -0.004 & 3 \\
DeltaCon & Euclidean & Sym & 0.154 & -0.116 & 3 \\
DeltaCon & Cosine & DTW & 0.214 & -0.099 & 3 \\
DeltaCon & Euclidean & KL & 0.310 & -0.345 & 3 \\
DeltaCon & Cosine & KL & 0.554 & -0.045 & 3 \\
DeltaCon & Cosine & Pearson & 0.157 & 0.331 & 2 \\
DeltaCon & Cosine & Spearman & 0.047 & 0.290 & 2 \\
DeltaCon & Cosine & Xcorr & 0.111 & 0.276 & 2 \\
DeltaCon & Euclidean & Xcorr & 0.116 & 0.284 & 2 \\
DeltaCon & Euclidean & Spearman & 0.048 & 0.287 & 2 \\
DeltaCon & Euclidean & Pearson & 0.161 & 0.338 & 2 \\
Spectral & Euclidean & Pearson & 0.179 & 0.284 & 1 \\
Spectral & Euclidean & Xcorr & 0.102 & 0.201 & 1 \\
Spectral & Cosine & Xcorr & 0.092 & 0.197 & 1 \\
Spectral & Cosine & Spearman & 0.025 & 0.157 & 1 \\
Spectral & Cosine & Pearson & 0.176 & 0.287 & 1 \\
Spectral & Euclidean & Spearman & 0.022 & 0.155 & 1 \\
Spectral & Cosine & KL & -0.288 & 0.106 & 0 \\
Spectral & Cosine & Sym & -0.213 & 0.099 & 0 \\
Spectral & Euclidean & DTW & -0.153 & 0.117 & 0 \\
Spectral & Euclidean & KL & -0.233 & 0.145 & 0 \\
Spectral & Euclidean & Gauss & -0.197 & 0.399 & 0 \\
Spectral & Cosine & Gauss & -0.546 & 0.316 & 0 \\
Spectral & Cosine & DTW & -0.214 & 0.099 & 0 \\
Spectral & Euclidean & Sym & -0.153 & 0.117 & 0 \\
\bottomrule
\end{longtable}

%% file: integrity_2merge100_aligned.tex
\begin{tabular}{lllrrrrrrrrr}
\toprule
\multicolumn{3}{c}{}%
& \multicolumn{2}{c}{\textbf{Non-learning-based}}%
& \multicolumn{3}{c}{\textbf{Static}}%
& \multicolumn{4}{c}{\textbf{Dynamic}}\\
\cmidrule(lr){4-5}\cmidrule(lr){6-8}\cmidrule(lr){9-12}
$\Delta G$ & $\Delta R$ & kernel & IPP & UASE & GAT & AE & AE-Shuffled & DynAE & DynRNN & DynAERNN & STConv \\
\midrule
DeltaCon & Cosine & Sym & 0.267 & \cellcolor{gold}0.601 & 0.249 & \cellcolor{bronze}0.338 & \cellcolor{silver}0.357 & 0.253 & 0.228 & 0.228 & 0.233 \\
DeltaCon & Cosine & Gauss & 0.0 & \cellcolor{gold}0.041 & 0.0 & \cellcolor{bronze}0.000 & \cellcolor{silver}0.000 & 0.0 & 0.0 & 0.0 & 0.0 \\
DeltaCon & Cosine & KL & 0.0 & \cellcolor{gold}0.659 & 0.005 & \cellcolor{bronze}0.006 & \cellcolor{silver}0.037 & 0.0 & 0.0 & 0.0 & 0.0 \\
DeltaCon & Cosine & Pearson & \cellcolor{bronze}0.798 & 0.637 & 0.471 & \cellcolor{gold}0.847 & 0.581 & \cellcolor{silver}0.799 & 0.5 & 0.644 & 0.605 \\
DeltaCon & Cosine & Spearman & 0.74 & 0.586 & 0.459 & \cellcolor{gold}0.807 & 0.575 & \cellcolor{silver}0.792 & 0.5 & \cellcolor{bronze}0.770 & 0.749 \\
DeltaCon & Cosine & Xcorr & \cellcolor{silver}0.807 & 0.674 & 0.495 & \cellcolor{gold}0.853 & 0.613 & \cellcolor{bronze}0.802 & 0.59 & 0.66 & 0.755 \\
DeltaCon & Cosine & DTW & 0.267 & \cellcolor{gold}0.601 & 0.249 & \cellcolor{bronze}0.338 & \cellcolor{silver}0.357 & 0.253 & 0.228 & 0.228 & 0.233 \\
DeltaCon & Euclidean & Sym & 0.365 & \cellcolor{gold}0.658 & 0.282 & \cellcolor{bronze}0.454 & \cellcolor{silver}0.477 & 0.336 & 0.228 & 0.238 & 0.241 \\
DeltaCon & Euclidean & Gauss & 0.0 & \cellcolor{gold}0.091 & 0.0 & \cellcolor{bronze}0.002 & \cellcolor{silver}0.004 & 0.0 & 0.0 & 0.0 & 0.0 \\
DeltaCon & Euclidean & KL & 0.034 & \cellcolor{gold}0.749 & 0.025 & \cellcolor{bronze}0.313 & \cellcolor{silver}0.387 & 0.001 & 0.0 & 0.0 & 0.0 \\
DeltaCon & Euclidean & Pearson & \cellcolor{bronze}0.803 & 0.638 & 0.46 & \cellcolor{gold}0.849 & 0.579 & \cellcolor{silver}0.813 & 0.509 & 0.697 & 0.623 \\
DeltaCon & Euclidean & Spearman & 0.739 & 0.587 & 0.476 & \cellcolor{gold}0.809 & 0.576 & \cellcolor{bronze}0.792 & 0.581 & 0.77 & \cellcolor{silver}0.794 \\
DeltaCon & Euclidean & Xcorr & \cellcolor{bronze}0.811 & 0.676 & 0.472 & \cellcolor{gold}0.853 & 0.612 & \cellcolor{silver}0.813 & 0.595 & 0.721 & 0.806 \\
DeltaCon & Euclidean & DTW & 0.365 & \cellcolor{gold}0.658 & 0.282 & \cellcolor{bronze}0.454 & \cellcolor{silver}0.477 & 0.336 & 0.228 & 0.238 & 0.241 \\
Spectral & Cosine & Sym & 0.963 & 0.628 & 0.979 & 0.891 & 0.872 & 0.977 & \cellcolor{silver}0.999 & \cellcolor{gold}0.999 & \cellcolor{bronze}0.996 \\
Spectral & Cosine & Gauss & 0.967 & 0.056 & 0.967 & 0.75 & 0.691 & 0.987 & \cellcolor{silver}1.000 & \cellcolor{gold}1.000 & \cellcolor{bronze}0.994 \\
Spectral & Cosine & KL & 0.966 & 0.54 & 0.98 & 0.889 & 0.869 & 0.98 & \cellcolor{bronze}0.985 & \cellcolor{gold}0.998 & \cellcolor{silver}0.997 \\
Spectral & Cosine & Pearson & \cellcolor{bronze}0.660 & 0.498 & 0.46 & \cellcolor{silver}0.752 & 0.535 & \cellcolor{gold}0.754 & 0.5 & 0.6 & 0.575 \\
Spectral & Cosine & Spearman & 0.652 & 0.536 & 0.476 & \cellcolor{gold}0.750 & 0.553 & \cellcolor{silver}0.732 & 0.5 & \cellcolor{bronze}0.699 & 0.698 \\
Spectral & Cosine & Xcorr & 0.669 & 0.523 & 0.478 & \cellcolor{bronze}0.752 & 0.608 & \cellcolor{silver}0.758 & 0.571 & \cellcolor{gold}0.778 & 0.693 \\
Spectral & Cosine & DTW & 0.963 & 0.628 & 0.98 & 0.891 & 0.872 & 0.977 & \cellcolor{silver}0.999 & \cellcolor{gold}0.999 & \cellcolor{bronze}0.997 \\
Spectral & Euclidean & Sym & 0.864 & 0.571 & 0.947 & 0.776 & 0.752 & 0.893 & \cellcolor{gold}0.999 & \cellcolor{silver}0.991 & \cellcolor{bronze}0.988 \\
Spectral & Euclidean & Gauss & 0.661 & 0.019 & 0.889 & 0.346 & 0.263 & 0.773 & \cellcolor{gold}1.000 & \cellcolor{silver}0.996 & \cellcolor{bronze}0.992 \\
Spectral & Euclidean & KL & 0.859 & 0.447 & 0.947 & 0.75 & 0.721 & 0.893 & \cellcolor{bronze}0.987 & \cellcolor{gold}0.993 & \cellcolor{silver}0.991 \\
Spectral & Euclidean & Pearson & \cellcolor{bronze}0.665 & 0.5 & 0.455 & \cellcolor{silver}0.747 & 0.536 & \cellcolor{gold}0.753 & 0.488 & 0.642 & 0.585 \\
Spectral & Euclidean & Spearman & 0.652 & 0.535 & 0.492 & \cellcolor{gold}0.752 & 0.555 & \cellcolor{silver}0.731 & 0.54 & 0.698 & \cellcolor{bronze}0.711 \\
Spectral & Euclidean & Xcorr & 0.673 & 0.526 & 0.476 & \cellcolor{bronze}0.747 & 0.612 & \cellcolor{silver}0.755 & 0.579 & \cellcolor{gold}0.810 & 0.716 \\
Spectral & Euclidean & DTW & 0.864 & 0.571 & 0.947 & 0.776 & 0.752 & 0.893 & \cellcolor{gold}0.999 & \cellcolor{silver}0.992 & \cellcolor{bronze}0.988 \\
Euclidean & Cosine & Sym & 0.506 & \cellcolor{gold}0.840 & 0.488 & \cellcolor{bronze}0.577 & \cellcolor{silver}0.596 & 0.491 & 0.467 & 0.467 & 0.472 \\
Euclidean & Cosine & Gauss & 0.006 & \cellcolor{gold}0.567 & 0.007 & \cellcolor{bronze}0.029 & \cellcolor{silver}0.037 & 0.006 & 0.003 & 0.003 & 0.007 \\
Euclidean & Cosine & KL & 0.023 & \cellcolor{gold}0.947 & 0.068 & \cellcolor{bronze}0.406 & \cellcolor{silver}0.518 & 0.016 & 0.0 & 0.0 & 0.009 \\
Euclidean & Cosine & Pearson & \cellcolor{silver}0.865 & \cellcolor{gold}0.996 & 0.482 & \cellcolor{bronze}0.667 & 0.595 & 0.506 & 0.5 & 0.301 & 0.396 \\
Euclidean & Cosine & Spearman & \cellcolor{silver}0.849 & \cellcolor{gold}0.992 & 0.411 & 0.562 & \cellcolor{bronze}0.663 & 0.516 & 0.5 & 0.471 & 0.6 \\
Euclidean & Cosine & Xcorr & \cellcolor{silver}0.918 & \cellcolor{gold}0.998 & 0.507 & \cellcolor{bronze}0.728 & 0.607 & 0.533 & 0.546 & 0.337 & 0.586 \\
Euclidean & Cosine & DTW & 0.506 & \cellcolor{gold}0.840 & 0.488 & \cellcolor{bronze}0.577 & \cellcolor{silver}0.596 & 0.491 & 0.467 & 0.467 & 0.472 \\
Euclidean & Euclidean & Sym & 0.604 & \cellcolor{gold}0.897 & 0.521 & \cellcolor{bronze}0.692 & \cellcolor{silver}0.716 & 0.575 & 0.467 & 0.477 & 0.48 \\
Euclidean & Euclidean & Gauss & 0.037 & \cellcolor{gold}0.783 & 0.019 & \cellcolor{bronze}0.150 & \cellcolor{silver}0.192 & 0.027 & 0.003 & 0.005 & 0.007 \\
Euclidean & Euclidean & KL & 0.56 & \cellcolor{gold}0.977 & 0.18 & \cellcolor{bronze}0.760 & \cellcolor{silver}0.808 & 0.443 & 0.0 & 0.013 & 0.009 \\
Euclidean & Euclidean & Pearson & \cellcolor{silver}0.866 & \cellcolor{gold}0.995 & 0.45 & \cellcolor{bronze}0.699 & 0.581 & 0.552 & 0.522 & 0.318 & 0.411 \\
Euclidean & Euclidean & Spearman & \cellcolor{silver}0.849 & \cellcolor{gold}0.992 & 0.423 & 0.562 & \cellcolor{bronze}0.661 & 0.516 & 0.388 & 0.474 & 0.654 \\
Euclidean & Euclidean & Xcorr & \cellcolor{silver}0.919 & \cellcolor{gold}0.998 & 0.475 & \cellcolor{bronze}0.754 & 0.594 & 0.575 & 0.533 & 0.367 & 0.67 \\
Euclidean & Euclidean & DTW & 0.604 & \cellcolor{gold}0.912 & 0.521 & \cellcolor{bronze}0.692 & \cellcolor{silver}0.716 & 0.575 & 0.467 & 0.477 & 0.48 \\
\bottomrule
\end{tabular}

%% file: integrity_3move100_aligned.tex
\begin{tabular}{lllrrrrrrrrr}
\toprule
\multicolumn{3}{c}{}%
& \multicolumn{2}{c}{\textbf{Non-learning-based}}%
& \multicolumn{3}{c}{\textbf{Static}}%
& \multicolumn{4}{c}{\textbf{Dynamic}}\\
\cmidrule(lr){4-5}\cmidrule(lr){6-8}\cmidrule(lr){9-12}
$\Delta G$ & $\Delta R$ & kernel & IPP & UASE & GAT & AE & AE-Shuffled & DynAE & DynRNN & DynAERNN & STConv \\
\midrule
DeltaCon & Cosine & Sym & 0.236 & \cellcolor{gold}0.516 & 0.249 & \cellcolor{bronze}0.345 & \cellcolor{silver}0.348 & 0.272 & 0.204 & 0.207 & 0.217 \\
DeltaCon & Cosine & Gauss & 0.0 & \cellcolor{gold}0.006 & \cellcolor{silver}0.002 & 0.0 & \cellcolor{bronze}0.000 & 0.0 & 0.0 & 0.0 & 0.0 \\
DeltaCon & Cosine & KL & 0.0 & \cellcolor{gold}0.499 & \cellcolor{silver}0.038 & 0.001 & \cellcolor{bronze}0.013 & 0.0 & 0.0 & 0.0 & 0.004 \\
DeltaCon & Cosine & Pearson & \cellcolor{gold}0.884 & 0.506 & 0.474 & \cellcolor{silver}0.775 & 0.416 & \cellcolor{bronze}0.740 & 0.5 & 0.673 & 0.546 \\
DeltaCon & Cosine & Spearman & \cellcolor{gold}0.815 & 0.246 & 0.48 & \cellcolor{bronze}0.780 & 0.399 & \cellcolor{silver}0.794 & 0.5 & 0.592 & 0.6 \\
DeltaCon & Cosine & Xcorr & \cellcolor{gold}0.884 & 0.506 & 0.537 & \cellcolor{silver}0.775 & 0.52 & \cellcolor{bronze}0.766 & 0.614 & 0.673 & 0.644 \\
DeltaCon & Cosine & DTW & 0.236 & \cellcolor{gold}0.516 & 0.249 & \cellcolor{bronze}0.345 & \cellcolor{silver}0.348 & 0.272 & 0.204 & 0.207 & 0.217 \\
DeltaCon & Euclidean & Sym & 0.329 & \cellcolor{gold}0.597 & 0.288 & \cellcolor{bronze}0.466 & \cellcolor{silver}0.470 & 0.386 & 0.204 & 0.239 & 0.228 \\
DeltaCon & Euclidean & Gauss & 0.0 & \cellcolor{gold}0.030 & \cellcolor{bronze}0.002 & 0.002 & \cellcolor{silver}0.002 & 0.0 & 0.0 & 0.0 & 0.0 \\
DeltaCon & Euclidean & KL & 0.0 & \cellcolor{gold}0.657 & 0.077 & \cellcolor{bronze}0.376 & \cellcolor{silver}0.386 & 0.112 & 0.0 & 0.0 & 0.003 \\
DeltaCon & Euclidean & Pearson & \cellcolor{gold}0.895 & 0.52 & 0.464 & \cellcolor{silver}0.779 & 0.416 & \cellcolor{bronze}0.745 & 0.47 & 0.644 & 0.584 \\
DeltaCon & Euclidean & Spearman & \cellcolor{gold}0.813 & 0.252 & 0.485 & \cellcolor{bronze}0.780 & 0.396 & \cellcolor{silver}0.795 & 0.455 & 0.59 & 0.75 \\
DeltaCon & Euclidean & Xcorr & \cellcolor{gold}0.895 & 0.52 & 0.54 & \cellcolor{silver}0.779 & 0.516 & \cellcolor{bronze}0.770 & 0.599 & 0.644 & 0.668 \\
DeltaCon & Euclidean & DTW & 0.329 & \cellcolor{gold}0.597 & 0.288 & \cellcolor{bronze}0.466 & \cellcolor{silver}0.470 & 0.386 & 0.204 & 0.239 & 0.228 \\
Spectral & Cosine & Sym & 0.969 & 0.69 & 0.954 & 0.861 & 0.858 & 0.933 & \cellcolor{gold}0.998 & \cellcolor{silver}0.997 & \cellcolor{bronze}0.987 \\
Spectral & Cosine & Gauss & 0.979 & 0.119 & 0.9 & 0.651 & 0.636 & 0.902 & \cellcolor{silver}0.997 & \cellcolor{gold}0.998 & \cellcolor{bronze}0.987 \\
Spectral & Cosine & KL & 0.973 & 0.636 & 0.945 & 0.856 & 0.852 & 0.935 & \cellcolor{bronze}0.978 & \cellcolor{gold}0.997 & \cellcolor{silver}0.989 \\
Spectral & Cosine & Pearson & \cellcolor{gold}0.975 & \cellcolor{silver}0.781 & 0.476 & 0.572 & 0.487 & 0.523 & 0.5 & \cellcolor{bronze}0.729 & 0.514 \\
Spectral & Cosine & Spearman & \cellcolor{gold}0.664 & 0.395 & 0.482 & \cellcolor{silver}0.646 & 0.479 & \cellcolor{bronze}0.645 & 0.5 & 0.532 & 0.557 \\
Spectral & Cosine & Xcorr & \cellcolor{gold}0.975 & \cellcolor{silver}0.781 & 0.564 & 0.572 & 0.575 & 0.551 & 0.649 & \cellcolor{bronze}0.729 & 0.578 \\
Spectral & Cosine & DTW & 0.969 & 0.69 & 0.96 & 0.862 & 0.859 & 0.933 & \cellcolor{gold}0.998 & \cellcolor{silver}0.997 & \cellcolor{bronze}0.987 \\
Spectral & Euclidean & Sym & 0.877 & 0.609 & 0.916 & 0.74 & 0.736 & 0.82 & \cellcolor{gold}0.998 & \cellcolor{bronze}0.966 & \cellcolor{silver}0.977 \\
Spectral & Euclidean & Gauss & 0.714 & 0.034 & 0.801 & 0.225 & 0.215 & 0.488 & \cellcolor{gold}0.997 & \cellcolor{bronze}0.970 & \cellcolor{silver}0.982 \\
Spectral & Euclidean & KL & 0.874 & 0.512 & 0.907 & 0.706 & 0.7 & 0.807 & \cellcolor{gold}0.982 & \cellcolor{bronze}0.970 & \cellcolor{silver}0.980 \\
Spectral & Euclidean & Pearson & \cellcolor{gold}0.956 & \cellcolor{silver}0.791 & 0.468 & 0.57 & 0.488 & 0.526 & 0.473 & \cellcolor{bronze}0.647 & 0.552 \\
Spectral & Euclidean & Spearman & \cellcolor{gold}0.664 & 0.395 & 0.485 & \cellcolor{silver}0.647 & 0.478 & \cellcolor{bronze}0.643 & 0.511 & 0.537 & 0.633 \\
Spectral & Euclidean & Xcorr & \cellcolor{gold}0.956 & \cellcolor{silver}0.791 & 0.591 & 0.57 & 0.571 & 0.552 & 0.638 & \cellcolor{bronze}0.647 & 0.552 \\
Spectral & Euclidean & DTW & 0.877 & 0.609 & 0.922 & 0.74 & 0.736 & 0.82 & \cellcolor{gold}0.998 & \cellcolor{bronze}0.966 & \cellcolor{silver}0.977 \\
Euclidean & Cosine & Sym & 0.672 & \cellcolor{gold}0.953 & 0.682 & \cellcolor{bronze}0.782 & \cellcolor{silver}0.785 & 0.709 & 0.64 & 0.644 & 0.654 \\
Euclidean & Cosine & Gauss & 0.095 & \cellcolor{gold}0.948 & 0.151 & \cellcolor{bronze}0.354 & \cellcolor{silver}0.364 & 0.159 & 0.059 & 0.062 & 0.078 \\
Euclidean & Cosine & KL & 0.393 & \cellcolor{gold}0.994 & 0.234 & \cellcolor{bronze}0.848 & \cellcolor{silver}0.855 & 0.637 & 0.0 & 0.002 & 0.041 \\
Euclidean & Cosine & Pearson & \cellcolor{gold}0.919 & \cellcolor{silver}0.870 & 0.468 & 0.445 & 0.51 & 0.402 & 0.5 & \cellcolor{bronze}0.716 & 0.491 \\
Euclidean & Cosine & Spearman & 0.23 & \cellcolor{gold}0.929 & 0.499 & 0.146 & \cellcolor{silver}0.577 & 0.123 & \cellcolor{bronze}0.500 & 0.46 & 0.391 \\
Euclidean & Cosine & Xcorr & \cellcolor{gold}0.919 & \cellcolor{silver}0.870 & 0.562 & 0.445 & 0.592 & 0.43 & 0.631 & \cellcolor{bronze}0.716 & 0.53 \\
Euclidean & Cosine & DTW & 0.672 & \cellcolor{gold}0.953 & 0.682 & \cellcolor{bronze}0.782 & \cellcolor{silver}0.785 & 0.709 & 0.64 & 0.644 & 0.654 \\
Euclidean & Euclidean & Sym & 0.766 & \cellcolor{gold}0.962 & 0.72 & \cellcolor{bronze}0.903 & \cellcolor{silver}0.907 & 0.823 & 0.64 & 0.676 & 0.665 \\
Euclidean & Euclidean & Gauss & 0.3 & \cellcolor{gold}0.965 & 0.245 & \cellcolor{bronze}0.808 & \cellcolor{silver}0.823 & 0.505 & 0.059 & 0.101 & 0.089 \\
Euclidean & Euclidean & KL & 0.82 & \cellcolor{gold}0.996 & 0.431 & \cellcolor{bronze}0.975 & \cellcolor{silver}0.977 & 0.909 & 0.0 & 0.386 & 0.272 \\
Euclidean & Euclidean & Pearson & \cellcolor{gold}0.890 & \cellcolor{silver}0.874 & 0.467 & 0.442 & 0.512 & 0.404 & 0.459 & \cellcolor{bronze}0.627 & 0.52 \\
Euclidean & Euclidean & Spearman & 0.232 & \cellcolor{gold}0.924 & 0.48 & 0.148 & \cellcolor{silver}0.579 & 0.121 & \cellcolor{bronze}0.488 & 0.454 & 0.189 \\
Euclidean & Euclidean & Xcorr & \cellcolor{gold}0.890 & \cellcolor{silver}0.874 & 0.596 & 0.442 & 0.589 & 0.43 & \cellcolor{bronze}0.636 & 0.627 & 0.52 \\
Euclidean & Euclidean & DTW & 0.766 & \cellcolor{gold}0.962 & 0.72 & \cellcolor{bronze}0.903 & \cellcolor{silver}0.907 & 0.823 & 0.64 & 0.676 & 0.665 \\
\bottomrule
\end{tabular}

%% file: integrity_periodic100_aligned.tex
\begin{tabular}{lllrrrrrrrrr}
\toprule
\multicolumn{3}{c}{}%
& \multicolumn{2}{c}{\textbf{Non-learning-based}}%
& \multicolumn{3}{c}{\textbf{Static}}%
& \multicolumn{4}{c}{\textbf{Dynamic}}\\
\cmidrule(lr){4-5}\cmidrule(lr){6-8}\cmidrule(lr){9-12}
$\Delta G$ & $\Delta R$ & kernel & IPP & UASE & GAT & AE & AE-Shuffled & DynAE & DynRNN & DynAERNN & STConv \\
\midrule
DeltaCon & Cosine & Sym & 0.272 & \cellcolor{gold}0.596 & 0.212 & \cellcolor{bronze}0.362 & \cellcolor{silver}0.371 & 0.269 & 0.189 & 0.196 & 0.205 \\
DeltaCon & Cosine & Gauss & 0.0 & \cellcolor{gold}0.041 & \cellcolor{silver}0.000 & 0.0 & \cellcolor{bronze}0.000 & 0.0 & 0.0 & 0.0 & 0.0 \\
DeltaCon & Cosine & KL & 0.0 & \cellcolor{gold}0.641 & 0.012 & \cellcolor{bronze}0.034 & \cellcolor{silver}0.086 & 0.0 & 0.0 & 0.0 & 0.003 \\
DeltaCon & Cosine & Pearson & \cellcolor{silver}0.896 & \cellcolor{gold}0.911 & 0.541 & 0.883 & 0.408 & \cellcolor{bronze}0.885 & 0.5 & 0.383 & 0.588 \\
DeltaCon & Cosine & Spearman & \cellcolor{gold}0.919 & \cellcolor{silver}0.915 & 0.556 & \cellcolor{bronze}0.906 & 0.441 & 0.902 & 0.5 & 0.538 & 0.859 \\
DeltaCon & Cosine & Xcorr & \cellcolor{silver}0.896 & \cellcolor{gold}0.911 & 0.63 & 0.883 & 0.561 & \cellcolor{bronze}0.885 & 0.599 & 0.517 & 0.795 \\
DeltaCon & Cosine & DTW & 0.272 & \cellcolor{gold}0.596 & 0.212 & \cellcolor{bronze}0.362 & \cellcolor{silver}0.371 & 0.269 & 0.189 & 0.196 & 0.205 \\
DeltaCon & Euclidean & Sym & 0.386 & \cellcolor{gold}0.636 & 0.242 & \cellcolor{bronze}0.480 & \cellcolor{silver}0.488 & 0.385 & 0.189 & 0.235 & 0.234 \\
DeltaCon & Euclidean & Gauss & 0.0 & \cellcolor{gold}0.057 & 0.0 & \cellcolor{bronze}0.003 & \cellcolor{silver}0.004 & 0.0 & 0.0 & 0.0 & 0.0 \\
DeltaCon & Euclidean & KL & 0.122 & \cellcolor{gold}0.716 & 0.03 & \cellcolor{bronze}0.417 & \cellcolor{silver}0.433 & 0.115 & 0.0 & 0.003 & 0.001 \\
DeltaCon & Euclidean & Pearson & \cellcolor{silver}0.905 & \cellcolor{gold}0.913 & 0.524 & 0.887 & 0.412 & \cellcolor{bronze}0.891 & 0.52 & 0.433 & 0.668 \\
DeltaCon & Euclidean & Spearman & \cellcolor{gold}0.919 & \cellcolor{silver}0.915 & 0.579 & \cellcolor{bronze}0.905 & 0.441 & 0.902 & 0.516 & 0.531 & 0.878 \\
DeltaCon & Euclidean & Xcorr & \cellcolor{silver}0.905 & \cellcolor{gold}0.913 & 0.617 & 0.887 & 0.562 & \cellcolor{bronze}0.891 & 0.625 & 0.534 & 0.783 \\
DeltaCon & Euclidean & DTW & 0.386 & \cellcolor{gold}0.636 & 0.242 & \cellcolor{bronze}0.480 & \cellcolor{silver}0.488 & 0.385 & 0.189 & 0.235 & 0.234 \\
Spectral & Cosine & Sym & 0.931 & 0.607 & 0.971 & 0.841 & 0.832 & 0.934 & \cellcolor{bronze}0.987 & \cellcolor{silver}0.987 & \cellcolor{gold}0.989 \\
Spectral & Cosine & Gauss & 0.878 & 0.063 & 0.955 & 0.57 & 0.536 & 0.901 & \cellcolor{gold}0.993 & \cellcolor{silver}0.993 & \cellcolor{bronze}0.990 \\
Spectral & Cosine & KL & 0.947 & 0.53 & \cellcolor{bronze}0.963 & 0.855 & 0.844 & 0.951 & 0.822 & \cellcolor{silver}0.980 & \cellcolor{gold}0.994 \\
Spectral & Cosine & Pearson & \cellcolor{bronze}0.655 & \cellcolor{silver}0.680 & 0.601 & 0.649 & 0.439 & \cellcolor{gold}0.697 & 0.5 & 0.429 & 0.572 \\
Spectral & Cosine & Spearman & \cellcolor{bronze}0.617 & 0.606 & 0.513 & 0.595 & 0.449 & \cellcolor{silver}0.631 & 0.5 & 0.472 & \cellcolor{gold}0.645 \\
Spectral & Cosine & Xcorr & 0.655 & \cellcolor{silver}0.680 & 0.601 & 0.649 & 0.493 & \cellcolor{gold}0.697 & 0.556 & 0.473 & \cellcolor{bronze}0.667 \\
Spectral & Cosine & DTW & 0.933 & 0.607 & 0.978 & 0.841 & 0.832 & 0.934 & \cellcolor{bronze}0.987 & \cellcolor{gold}0.993 & \cellcolor{silver}0.991 \\
Spectral & Euclidean & Sym & 0.817 & 0.567 & 0.955 & 0.722 & 0.715 & 0.818 & \cellcolor{gold}0.987 & \cellcolor{bronze}0.965 & \cellcolor{silver}0.968 \\
Spectral & Euclidean & Gauss & 0.486 & 0.021 & 0.904 & 0.182 & 0.169 & 0.481 & \cellcolor{gold}0.993 & \cellcolor{bronze}0.953 & \cellcolor{silver}0.972 \\
Spectral & Euclidean & KL & 0.827 & 0.468 & \cellcolor{bronze}0.961 & 0.71 & 0.699 & 0.829 & 0.843 & \cellcolor{silver}0.973 & \cellcolor{gold}0.979 \\
Spectral & Euclidean & Pearson & \cellcolor{bronze}0.669 & \cellcolor{silver}0.685 & 0.563 & 0.654 & 0.441 & \cellcolor{gold}0.701 & 0.541 & 0.443 & 0.6 \\
Spectral & Euclidean & Spearman & \cellcolor{silver}0.616 & 0.603 & 0.528 & 0.595 & 0.448 & \cellcolor{gold}0.631 & 0.499 & 0.465 & \cellcolor{bronze}0.606 \\
Spectral & Euclidean & Xcorr & \cellcolor{bronze}0.669 & \cellcolor{silver}0.685 & 0.563 & 0.654 & 0.492 & \cellcolor{gold}0.701 & 0.572 & 0.453 & 0.626 \\
Spectral & Euclidean & DTW & 0.817 & 0.567 & 0.96 & 0.722 & 0.715 & 0.818 & \cellcolor{gold}0.987 & \cellcolor{silver}0.975 & \cellcolor{bronze}0.971 \\
Euclidean & Cosine & Sym & 0.502 & \cellcolor{gold}0.826 & 0.443 & \cellcolor{bronze}0.592 & \cellcolor{silver}0.601 & 0.499 & 0.42 & 0.426 & 0.436 \\
Euclidean & Cosine & Gauss & 0.008 & \cellcolor{gold}0.511 & 0.011 & \cellcolor{bronze}0.050 & \cellcolor{silver}0.078 & 0.01 & 0.003 & 0.004 & 0.004 \\
Euclidean & Cosine & KL & 0.158 & \cellcolor{gold}0.936 & 0.059 & \cellcolor{bronze}0.569 & \cellcolor{silver}0.579 & 0.168 & 0.0 & 0.012 & 0.006 \\
Euclidean & Cosine & Pearson & \cellcolor{silver}0.996 & \cellcolor{gold}0.999 & 0.552 & 0.959 & 0.42 & \cellcolor{bronze}0.961 & 0.5 & 0.377 & 0.633 \\
Euclidean & Cosine & Spearman & \cellcolor{gold}0.987 & \cellcolor{silver}0.983 & 0.603 & \cellcolor{bronze}0.960 & 0.444 & 0.955 & 0.5 & 0.524 & 0.88 \\
Euclidean & Cosine & Xcorr & \cellcolor{silver}0.996 & \cellcolor{gold}0.999 & 0.636 & 0.959 & 0.562 & \cellcolor{bronze}0.961 & 0.609 & 0.543 & 0.791 \\
Euclidean & Cosine & DTW & 0.502 & \cellcolor{gold}0.854 & 0.447 & \cellcolor{bronze}0.592 & \cellcolor{silver}0.601 & 0.499 & 0.42 & 0.426 & 0.436 \\
Euclidean & Euclidean & Sym & 0.616 & \cellcolor{gold}0.866 & 0.473 & \cellcolor{bronze}0.711 & \cellcolor{silver}0.718 & 0.615 & 0.42 & 0.466 & 0.465 \\
Euclidean & Euclidean & Gauss & 0.049 & \cellcolor{gold}0.664 & 0.022 & \cellcolor{bronze}0.216 & \cellcolor{silver}0.259 & 0.062 & 0.003 & 0.013 & 0.007 \\
Euclidean & Euclidean & KL & 0.635 & \cellcolor{gold}0.958 & 0.111 & \cellcolor{bronze}0.806 & \cellcolor{silver}0.808 & 0.633 & 0.0 & 0.089 & 0.046 \\
Euclidean & Euclidean & Pearson & \cellcolor{gold}0.999 & \cellcolor{silver}0.998 & 0.542 & 0.961 & 0.425 & \cellcolor{bronze}0.965 & 0.556 & 0.427 & 0.738 \\
Euclidean & Euclidean & Spearman & \cellcolor{gold}0.987 & \cellcolor{silver}0.983 & 0.634 & \cellcolor{bronze}0.960 & 0.443 & 0.955 & 0.573 & 0.517 & 0.95 \\
Euclidean & Euclidean & Xcorr & \cellcolor{gold}0.999 & \cellcolor{silver}0.998 & 0.633 & 0.961 & 0.561 & \cellcolor{bronze}0.965 & 0.637 & 0.529 & 0.842 \\
Euclidean & Euclidean & DTW & 0.616 & \cellcolor{gold}0.886 & 0.473 & \cellcolor{bronze}0.711 & \cellcolor{silver}0.718 & 0.615 & 0.42 & 0.466 & 0.465 \\
\bottomrule
\end{tabular}

%% file: main.bbl
\begin{thebibliography}{10}

\bibitem{10.5555/3327546.3327592}
{\sc Abu-El-Haija, S., Perozzi, B., Al-Rfou, R., and Alemi, A.}
\newblock Watch your step: learning node embeddings via graph attention.
\newblock In {\em Proceedings of the 32nd International Conference on Neural Information Processing Systems\/} (Red Hook, NY, USA, 2018), NIPS'18, Curran Associates Inc., p.~9198–9208.

\bibitem{agarwal2021unified}
{\sc Agarwal, C., Lakkaraju, H., and Zitnik, M.}
\newblock Towards a unified framework for fair and stable graph representation learning, 2021.

\bibitem{10.1145/3308558.3313668}
{\sc Al-Rfou, R., Perozzi, B., and Zelle, D.}
\newblock Ddgk: Learning graph representations for deep divergence graph kernels.
\newblock In {\em The World Wide Web Conference\/} (New York, NY, USA, 2019), WWW '19, Association for Computing Machinery, p.~37–48.

\bibitem{alsayouri2018recs}
{\sc Al-Sayouri, S.~A., Koutra, D., Papalexakis, E.~E., and Lam, S.~S.}
\newblock Recs: Robust graph embedding using connection subgraphs, 2018.

\bibitem{antoniak-mimno-2018-evaluating}
{\sc Antoniak, M., and Mimno, D.}
\newblock Evaluating the stability of embedding-based word similarities.
\newblock {\em Transactions of the Association for Computational Linguistics 6\/} (2018), 107--119.

\bibitem{10.1145/3483595}
{\sc Barros, C. D.~T., Mendon\c{c}a, M. R.~F., Vieira, A.~B., and Ziviani, A.}
\newblock A survey on embedding dynamic graphs.

\bibitem{bechlerspeicher2025positiongraphlearninglose}
{\sc Bechler-Speicher, M., Finkelshtein, B., Frasca, F., Müller, L., Tönshoff, J., Siraudin, A., Zaverkin, V., Bronstein, M.~M., Niepert, M., Perozzi, B., Galkin, M., and Morris, C.}
\newblock Position: Graph learning will lose relevance due to poor benchmarks, 2025.

\bibitem{Belkin2001LaplacianEA}
{\sc Belkin, M., and Niyogi, P.}
\newblock Laplacian eigenmaps and spectral techniques for embedding and clustering.
\newblock In {\em Neural Information Processing Systems\/} (2001).

\bibitem{Belkin2003LaplacianEF}
{\sc Belkin, M., and Niyogi, P.}
\newblock Laplacian eigenmaps for dimensionality reduction and data representation.
\newblock {\em Neural Computation 15\/} (2003), 1373--1396.

\bibitem{2011snda.book..115B}
{\sc {Bhagat}, S., {Cormode}, G., and {Muthukrishnan}, S.}
\newblock {Node Classification in Social Networks}.
\newblock In {\em Social Network Data Analytics}, C.~C. {Aggarwal}, Ed. 2011, p.~115.

\bibitem{Bo2020StructuralDC}
{\sc Bo, D., Wang, X., Shi, C., Zhu, M., Lu, E., and Cui, P.}
\newblock Structural deep clustering network.
\newblock {\em Proceedings of The Web Conference 2020\/} (2020).

\bibitem{10.1145/3465336.3475098}
{\sc Borah, A., Barman, M.~P., and Awekar, A.}
\newblock Are word embedding methods stable and should we care about it?
\newblock In {\em Proceedings of the 32nd ACM Conference on Hypertext and Social Media\/} (New York, NY, USA, 2021), HT '21, Association for Computing Machinery, p.~45–55.

\bibitem{10.1145/3269206.3271740}
{\sc Chen, H., and Li, J.}
\newblock Exploiting structural and temporal evolution in dynamic link prediction.
\newblock In {\em Proceedings of the 27th ACM International Conference on Information and Knowledge Management\/} (New York, NY, USA, 2018), CIKM '18, Association for Computing Machinery, p.~427–436.

\bibitem{8392745}
{\sc Cui, P., Wang, X., Pei, J., and Zhu, W.}
\newblock A survey on network embedding.
\newblock {\em IEEE Transactions on Knowledge and Data Engineering 31}, 5 (2019), 833--852.

\bibitem{6252471}
{\sc da~Silva~Soares, P.~R., and Prudêncio, R. B.~C.}
\newblock Time series based link prediction.
\newblock In {\em The 2012 International Joint Conference on Neural Networks (IJCNN)\/} (2012), pp.~1--7.

\bibitem{davis2023simplepowerfulframeworkstable}
{\sc Davis, E., Gallagher, I., Lawson, D.~J., and Rubin-Delanchy, P.}
\newblock A simple and powerful framework for stable dynamic network embedding.

\bibitem{10.1007/s10994-023-06475-x}
{\sc Dileo, M., Zignani, M., and Gaito, S.}
\newblock Temporal graph learning for dynamic link prediction with text in online social networks.
\newblock {\em Mach. Learn. 113}, 4 (nov 2023), 2207–2226.

\bibitem{a58b9be8ce8a4c09a4d6f948ff3d5a53}
{\sc Dridi, A., Gaber, M., Azad, R., and Bhogal, J.}
\newblock Vec2dynamics: A temporal word embedding approach to exploring the dynamics of scientific keywords—machine learning as a case study.
\newblock {\em Big Data and Cognitive Computing 6}, 1 (Mar. 2022).
\newblock Publisher Copyright: {\textcopyright} 2022 by the authors. Licensee MDPI, Basel, Switzerland.

\bibitem{Faloutsos2013DELTACONAP}
{\sc Faloutsos, C., Koutra, D., and Vogelstein, J.~T.}
\newblock Deltacon: A principled massive-graph similarity function.
\newblock In {\em SDM\/} (2013).

\bibitem{10.5555/3540261.3541038}
{\sc Gallagher, I., Jones, A., and Rubin-Delanchy, P.}
\newblock Spectral embedding for dynamic networks with stability guarantees.
\newblock In {\em Proceedings of the 35th International Conference on Neural Information Processing Systems\/} (Red Hook, NY, USA, 2024), NIPS '21, Curran Associates Inc.

\bibitem{10.5555/3304222.3304235}
{\sc Gao, H., and Huang, H.}
\newblock Deep attributed network embedding.
\newblock In {\em Proceedings of the 27th International Joint Conference on Artificial Intelligence\/} (2018), IJCAI'18, AAAI Press, p.~3364–3370.

\bibitem{gong-etal-2020-enriching}
{\sc Gong, H., Bhat, S., and Viswanath, P.}
\newblock Enriching word embeddings with temporal and spatial information.
\newblock In {\em Proceedings of the 24th Conference on Computational Natural Language Learning\/} (Online, Nov. 2020), R.~Fern{\'a}ndez and T.~Linzen, Eds., Association for Computational Linguistics, pp.~1--11.

\bibitem{Goyal2018dyngraph2vecCN}
{\sc Goyal, P., Chhetri, S.~R., and Canedo, A.}
\newblock dyngraph2vec: Capturing network dynamics using dynamic graph representation learning.
\newblock {\em ArXiv abs/1809.02657\/} (2018).

\bibitem{goyal2018dyngem}
{\sc Goyal, P., Kamra, N., He, X., and Liu, Y.}
\newblock Dyngem: Deep embedding method for dynamic graphs.
\newblock {\em arXiv preprint arXiv:1805.11273\/} (2018).

\bibitem{doi:10.1080/23311983.2016.1171458}
{\sc Grandjean, M.}
\newblock A social network analysis of twitter: Mapping the digital humanities community.
\newblock {\em Cogent Arts \& Humanities 3}, 1 (2016), 1171458.

\bibitem{10.1145/2939672.2939754}
{\sc Grover, A., and Leskovec, J.}
\newblock node2vec: Scalable feature learning for networks.
\newblock In {\em Proceedings of the 22nd ACM SIGKDD International Conference on Knowledge Discovery and Data Mining\/} (New York, NY, USA, 2016), KDD '16, Association for Computing Machinery, p.~855–864.

\bibitem{DBLP:journals/snam/GursoyB22}
{\sc G{\"{u}}rsoy, F., and Badur, B.}
\newblock Investigating internal migration with network analysis and latent space representations: an application to turkey.
\newblock {\em Soc. Netw. Anal. Min. 12}, 1 (2022), 150.

\bibitem{alma9972283310208496}
{\sc Guzzi, P.~H., and Roy, S.}
\newblock {\em Biological network analysis : trends, approaches, graphical theory and algorithms}.
\newblock Elsevier, Amsterdam, Netherlands ;, 2020.

\bibitem{gürsoy2021alignment}
{\sc Gürsoy, F., Haddad, M., and Bothorel, C.}
\newblock Alignment and stability of embeddings: measurement and inference improvement, 2021.

\bibitem{10.5555/3294771.3294869}
{\sc Hamilton, W.~L., Ying, R., and Leskovec, J.}
\newblock Inductive representation learning on large graphs.
\newblock In {\em Proceedings of the 31st International Conference on Neural Information Processing Systems\/} (Red Hook, NY, USA, 2017), NIPS'17, Curran Associates Inc., p.~1025–1035.

\bibitem{Hamilton2017RepresentationLO}
{\sc Hamilton, W.~L., Ying, R., and Leskovec, J.}
\newblock Representation learning on graphs: Methods and applications.
\newblock {\em IEEE Data Eng. Bull. 40\/} (2017), 52--74.

\bibitem{hellrich-etal-2019-influence}
{\sc Hellrich, J., Kampe, B., and Hahn, U.}
\newblock The influence of down-sampling strategies on {SVD} word embedding stability.
\newblock In {\em Proceedings of the 3rd Workshop on Evaluating Vector Space Representations for {NLP}\/} (Minneapolis, USA, June 2019), Association for Computational Linguistics, pp.~18--26.

\bibitem{Holland1983StochasticBF}
{\sc Holland, P., Laskey, K.~B., and Leinhardt, S.}
\newblock Stochastic blockmodels: First steps.
\newblock {\em Social Networks 5\/} (1983), 109--137.

\bibitem{e3da81eedd1340ac81c098b22fe0df41}
{\sc Holme, P., and Saramaki, J.}
\newblock Temporal networks.
\newblock {\em PHYSICS REPORTS: REVIEW SECTION OF PHYSICS LETTERS 519}, 3 (2012), 97--125.

\bibitem{Holme_2012}
{\sc Holme, P., and Saramäki, J.}
\newblock Temporal networks.
\newblock {\em Physics Reports 519}, 3 (Oct. 2012), 97–125.

\bibitem{Jones2020TheMR}
{\sc Jones, A., and Rubin-Delanchy, P.}
\newblock The multilayer random dot product graph.
\newblock {\em ArXiv abs/2007.10455\/} (2020).

\bibitem{8917213}
{\sc Kang, Z., Xu, H., Hu, J., and Pei, X.}
\newblock Learning dynamic graph embedding for traffic flow forecasting: A graph self-attentive method.
\newblock In {\em 2019 IEEE Intelligent Transportation Systems Conference (ITSC)\/} (2019), pp.~2570--2576.

\bibitem{JMLR:v21:19-447}
{\sc Kazemi, S.~M., Goel, R., Jain, K., Kobyzev, I., Sethi, A., Forsyth, P., and Poupart, P.}
\newblock Representation learning for dynamic graphs: A survey.
\newblock {\em Journal of Machine Learning Research 21}, 70 (2020), 1--73.

\bibitem{10.1145/3633518}
{\sc Khoshraftar, S., and An, A.}
\newblock A survey on graph representation learning methods.
\newblock {\em ACM Trans. Intell. Syst. Technol. 15}, 1 (jan 2024).

\bibitem{2022IEEEA..10k1820K}
{\sc {Kim}, H., {Lee}, B.~S., {Shin}, W.-Y., and {Lim}, S.}
\newblock {Graph Anomaly Detection With Graph Neural Networks: Current Status and Challenges}.
\newblock {\em IEEE Access 10\/} (Jan. 2022), 111820--111829.

\bibitem{Kipf2016VariationalGA}
{\sc Kipf, T., and Welling, M.}
\newblock Variational graph auto-encoders.
\newblock {\em ArXiv abs/1611.07308\/} (2016).

\bibitem{kipf2017semisupervised}
{\sc Kipf, T.~N., and Welling, M.}
\newblock Semi-supervised classification with graph convolutional networks.
\newblock In {\em International Conference on Learning Representations\/} (2017).

\bibitem{klabunde2022prediction}
{\sc Klabunde, M., and Lemmerich, F.}
\newblock On the prediction instability of graph neural networks, 2022.

\bibitem{10.1145/2824443}
{\sc Koutra, D., Shah, N., Vogelstein, J.~T., Gallagher, B., and Faloutsos, C.}
\newblock Deltacon: Principled massive-graph similarity function with attribution.
\newblock {\em ACM Trans. Knowl. Discov. Data 10}, 3 (Feb. 2016).

\bibitem{9892658}
{\sc Krzywda, M., Łukasik, S., and Gandomi, A.~H.}
\newblock Graph neural networks in computer vision - architectures, datasets and common approaches.
\newblock In {\em 2022 International Joint Conference on Neural Networks (IJCNN)\/} (2022), pp.~1--10.

\bibitem{10.1111:j.1467-8659.2011.01898.x}
{\sc Landesberger, T.~v., Kuijper, A., Schreck, T., Kohlhammer, J., Wijk, J.~v., Fekete, J., and Fellner, D.~W.}
\newblock {Visual Analysis of Large Graphs: State-of-the-Art and Future Research Challenges}.
\newblock {\em Computer Graphics Forum\/} (2011).

\bibitem{Li2023AMG}
{\sc Li, B., and Nabavi, S.}
\newblock A multimodal graph neural network framework for cancer molecular subtype classification.
\newblock {\em BMC Bioinformatics 25\/} (2023).

\bibitem{Li2024GraphNN}
{\sc Li, H., Zhao, Y., Mao, Z., Qin, Y., Xiao, Z., Feng, J., Gu, Y., Ju, W., Luo, X., and Zhang, M.}
\newblock Graph neural networks in intelligent transportation systems: Advances, applications and trends.

\bibitem{Li2022}
{\sc Li, Y., Zhou, X., and Pan, M.}
\newblock {\em Graph Neural Networks in Urban Intelligence}.
\newblock Springer Nature Singapore, Singapore, 2022, pp.~579--593.

\bibitem{Mahdavi2019DynamicJV}
{\sc Mahdavi, S., Khoshraftar, S., and An, A.}
\newblock Dynamic joint variational graph autoencoders.
\newblock {\em ArXiv abs/1910.01963\/} (2019).

\bibitem{Manessi_2020}
{\sc Manessi, F., Rozza, A., and Manzo, M.}
\newblock Dynamic graph convolutional networks.
\newblock {\em Pattern Recognition 97\/} (Jan. 2020), 107000.

\bibitem{modell2023intensity}
{\sc Modell, A., Gallagher, I., Ceccherini, E., Whiteley, N., and Rubin-Delanchy, P.}
\newblock Intensity profile projection: A framework for continuous-time representation learning for dynamic networks.
\newblock In {\em Thirty-seventh Conference on Neural Information Processing Systems\/} (2023).

\bibitem{10.1093/acprof:oso/9780199206650.001.0001}
{\sc Newman, M.}
\newblock {\em {Networks: An Introduction}}.
\newblock Oxford University Press, 03 2010.

\bibitem{d69168735a23434b90720e92b752c533}
{\sc Ng, A., Jordan, M., and Weiss, Y.}
\newblock On spectral clustering: Analysis and an algorithm.
\newblock In {\em Advances in Neural Information Processing Systems 14 - Proceedings of the 2001 Conference, NIPS 2001\/} (2002), Advances in Neural Information Processing Systems, Neural information processing systems foundation.
\newblock 15th Annual Neural Information Processing Systems Conference, NIPS 2001 ; Conference date: 03-12-2001 Through 08-12-2001.

\bibitem{pensky2017spectral}
{\sc Pensky, M., and Zhang, T.}
\newblock Spectral clustering in the dynamic stochastic block model, 2017.

\bibitem{10.1145/2623330.2623732}
{\sc Perozzi, B., Al-Rfou, R., and Skiena, S.}
\newblock Deepwalk: online learning of social representations.
\newblock In {\em Proceedings of the 20th ACM SIGKDD International Conference on Knowledge Discovery and Data Mining\/} (New York, NY, USA, 2014), KDD '14, Association for Computing Machinery, p.~701–710.

\bibitem{Poursafaei2022TowardsBE}
{\sc Poursafaei, F., Huang, S., Pelrine, K., and Rabbany, R.}
\newblock Towards better evaluation for dynamic link prediction.
\newblock {\em ArXiv abs/2207.10128\/} (2022).

\bibitem{qian2024mdgnn}
{\sc Qian, H., Zhou, H., Zhao, Q., Chen, H., Yao, H., Wang, J., Liu, Z., Yu, F., Zhang, Z., and Zhou, J.}
\newblock Mdgnn: Multi-relational dynamic graph neural network for comprehensive and dynamic stock investment prediction.
\newblock In {\em Proceedings of the AAAI Conference on Artificial Intelligence\/} (2024), vol.~38, pp.~14642--14650.

\bibitem{quan2024clustering}
{\sc Quan, R., Wang, W., Ma, F., Fan, H., and Yang, Y.}
\newblock Clustering for protein representation learning.
\newblock In {\em Proceedings of the IEEE/CVF Conference on Computer Vision and Pattern Recognition\/} (2024), pp.~319--329.

\bibitem{doi:10.1126/science.290.5500.2323}
{\sc Roweis, S.~T., and Saul, L.~K.}
\newblock Nonlinear dimensionality reduction by locally linear embedding.
\newblock {\em Science 290}, 5500 (2000), 2323--2326.

\bibitem{4700287}
{\sc Scarselli, F., Gori, M., Tsoi, A.~C., Hagenbuchner, M., and Monfardini, G.}
\newblock The graph neural network model.
\newblock {\em IEEE Transactions on Neural Networks 20}, 1 (2009), 61--80.

\bibitem{schumacher2020effects}
{\sc Schumacher, T., Wolf, H., Ritzert, M., Lemmerich, F., Bachmann, J., Frantzen, F., Klabunde, M., Grohe, M., and Strohmaier, M.}
\newblock The effects of randomness on the stability of node embeddings, 2020.

\bibitem{shi2023geometric}
{\sc Shi, B., Morris, E., Shen, H., Du, W., and Sajjad, M.~H.}
\newblock Geometric instability of graph neural networks on large graphs.
\newblock In {\em The Second Learning on Graphs Conference\/} (2023).

\bibitem{9439502}
{\sc Skarding, J., Gabrys, B., and Musial, K.}
\newblock Foundations and modeling of dynamic networks using dynamic graph neural networks: A survey.
\newblock {\em IEEE Access 9\/} (2021), 79143--79168.

\bibitem{10.1145/3539618.3591674}
{\sc Tang, H., Wu, S., Xu, G., and Li, Q.}
\newblock Dynamic graph evolution learning for recommendation.
\newblock In {\em Proceedings of the 46th International ACM SIGIR Conference on Research and Development in Information Retrieval\/} (New York, NY, USA, 2023), SIGIR '23, Association for Computing Machinery, p.~1589–1598.

\bibitem{Tantardini2019ComparingMF}
{\sc Tantardini, M., Ieva, F., Tajoli, L., and Piccardi, C.}
\newblock Comparing methods for comparing networks.
\newblock {\em Scientific Reports 9\/} (2019).

\bibitem{49951}
{\sc Tsitsulin, A., Palowitch, J., and Perozzi, B.}
\newblock Graph clustering with graph neural networks.

\bibitem{veličković2018graph}
{\sc Veličković, P., Cucurull, G., Casanova, A., Romero, A., Liò, P., and Bengio, Y.}
\newblock Graph attention networks.
\newblock In {\em International Conference on Learning Representations\/} (2018).

\bibitem{9076342}
{\sc Wang, C., Rao, W., Guo, W., Wang, P., Liu, J., and Guan, X.}
\newblock Towards understanding the instability of network embedding.
\newblock {\em IEEE Transactions on Knowledge and Data Engineering 34}, 2 (2022), 927--941.

\bibitem{10.1145/2939672.2939753}
{\sc Wang, D., Cui, P., and Zhu, W.}
\newblock Structural deep network embedding.
\newblock In {\em Proceedings of the 22nd ACM SIGKDD International Conference on Knowledge Discovery and Data Mining\/} (New York, NY, USA, 2016), KDD '16, Association for Computing Machinery, p.~1225–1234.

\bibitem{wang2023scalableresourcemanagementdynamic}
{\sc Wang, X., Cheng, N., Fu, L., Quan, W., Sun, R., Hui, Y., Luan, T., and Shen, X.}
\newblock Scalable resource management for dynamic mec: An unsupervised link-output graph neural network approach, 2023.

\bibitem{west01}
{\sc West, D.~B.}
\newblock {\em Introduction to Graph Theory}, second~ed.
\newblock Prentice Hall, Upper Saddle River, N.J., 2001.

\bibitem{Xu2018HowPA}
{\sc Xu, K., Hu, W., Leskovec, J., and Jegelka, S.}
\newblock How powerful are graph neural networks?
\newblock {\em ArXiv abs/1810.00826\/} (2018).

\bibitem{10.1016/j.neucom.2021.03.138}
{\sc Xue, G., Zhong, M., Li, J., Chen, J., Zhai, C., and Kong, R.}
\newblock Dynamic network embedding survey.
\newblock {\em Neurocomput. 472}, C (feb 2022), 212–223.

\bibitem{10.1609/aaai.v33i01.33017370}
{\sc Yao, L., Mao, C., and Luo, Y.}
\newblock Graph convolutional networks for text classification.
\newblock In {\em Proceedings of the Thirty-Third AAAI Conference on Artificial Intelligence and Thirty-First Innovative Applications of Artificial Intelligence Conference and Ninth AAAI Symposium on Educational Advances in Artificial Intelligence\/} (2019), AAAI'19/IAAI'19/EAAI'19, AAAI Press.

\bibitem{9679093}
{\sc Zhang, X., Zhang, L., Jin, B., and Lu, X.}
\newblock A multi-view confidence-calibrated framework for fair and stable graph representation learning.
\newblock In {\em 2021 IEEE International Conference on Data Mining (ICDM)\/} (2021), pp.~1493--1498.

\bibitem{zheng2024improving}
{\sc Zheng, X., Wu, B., Zhang, A.~X., and Li, W.}
\newblock Improving robustness of gnn-based anomaly detection by graph adversarial training.
\newblock In {\em Proceedings of the 2024 Joint International Conference on Computational Linguistics, Language Resources and Evaluation (LREC-COLING 2024)\/} (2024), pp.~8902--8912.

\bibitem{zhou2020graph}
{\sc Zhou, J., Cui, G., Hu, S., Zhang, Z., Yang, C., Liu, Z., Wang, L., Li, C., and Sun, M.}
\newblock Graph neural networks: A review of methods and applications.
\newblock {\em AI open 1\/} (2020), 57--81.

\bibitem{Zhou2018GraphNN}
{\sc Zhou, J., Cui, G., Zhang, Z., Yang, C., Liu, Z., and Sun, M.}
\newblock Graph neural networks: A review of methods and applications.
\newblock {\em ArXiv abs/1812.08434\/} (2018).

\bibitem{7511675}
{\sc Zhu, L., Guo, D., Yin, J., Steeg, G.~V., and Galstyan, A.}
\newblock Scalable temporal latent space inference for link prediction in dynamic social networks.
\newblock {\em IEEE Transactions on Knowledge and Data Engineering 28}, 10 (2016), 2765--2777.

\end{thebibliography}
